\documentclass{nature}

\usepackage{upgreek}
\usepackage{helvet}
\usepackage{lineno}

\usepackage{setspace}
\usepackage{amsfonts}
\usepackage{amssymb}
\usepackage{amsmath}

\usepackage{float}
\usepackage{graphicx}

\usepackage[export]{adjustbox}

\usepackage[nomarkers,nolists]{endfloat}

\usepackage{lscape}
\usepackage{afterpage}

\usepackage{pdfpages}
\usepackage{float}

\usepackage{caption}

\newcounter{mybodyfigure}
\newcounter{myedfigure}

\newcommand{\stepbodyfigure}{\refstepcounter{mybodyfigure}}
\DeclareRobustCommand{\bodyfigure}[1]{\stepbodyfigure\label{#1}{\themybodyfigure}}
\newcommand{\bodyfigurelabel}[1]{\textbf{Fig.~\bodyfigure{#1}:}}

\makeatletter
\g@addto@macro\caption@prepareslc{%
\renewcommand{\stepbodyfigure}{\caption@l@stepcounter{mybodyfigure}}}
\makeatother

\newcommand{\stepedfigure}{\refstepcounter{myedfigure}}
\DeclareRobustCommand{\edfigure}[1]{\stepedfigure\label{#1}{\themyedfigure}}
\newcommand{\edfigurelabel}[1]{\textbf{Extended Data Fig.~\edfigure{#1}:}}

\makeatletter
\g@addto@macro\caption@prepareslc{%
\renewcommand{\stepedfigure}{\caption@l@stepcounter{myedfigure}}}
\makeatother

\usepackage[super,comma,sort&compress]{natbib}

\usepackage[hidelinks]{hyperref}
\usepackage{verbatim}

\title{Deep-Learning Investigation of Vibrational Raman Spectra for Plant-Stress Analysis}

\author{Anoop C. Patil$^{1*}$,  Benny Jian Rong Sng$^{1,2*}$, Yu-Wei Chang$^{3*}$, Joana B. Pereira$^{4}$, Chua Nam-Hai$^{1,2}$, Rajani Sarojam$^{1,2}$, Gajendra Pratap Singh$^{1+}$, In‑Cheol Jang$^{1,2,5+}$, and Giovanni Volpe$^{1,3,6+}$}

\begin{document}
\sloppy
\maketitle
\begin{affiliations}
 \item Disruptive \& Sustainable Technologies for Agricultural Precision, 1 CREATE way, Singapore-MIT Alliance for Research and Technology, Singapore 138602, Singapore.
 \item Temasek Life Sciences Laboratory, 1 Research Link, National University of Singapore, Singapore 117604, Singapore.
 \item Department of Physics, University of Gothenburg, Gothenburg 41296, Sweden.
 \item Department of Clinical Neuroscience, Karolinska Institute, Stockholm 17165, Sweden.
\item Department of Biological Sciences, National University of Singapore, Singapore 117543, Singapore.
\item Science for Life Laboratory, Department of Physics, University of Gothenburg, Gothenburg 41296, Sweden.
 
 \vspace{10mm}
 \item[] $^{*}$Authors contributed equally to this work
 \item[] $^{+}$To whom correspondence should be addressed: \\giovanni.volpe@physics.gu.se; jangi@tll.org.sg; gajendra@smart.mit.edu
\end{affiliations}

\pagebreak

\begin{abstract}
Detecting stress in plants is crucial for both open-farm and controlled-environment agriculture. Biomolecules within plants serve as key stress indicators, offering vital markers for continuous health monitoring and early disease detection. Raman spectroscopy provides a powerful, non-invasive means to quantify these biomolecules through their molecular vibrational signatures. However, traditional Raman analysis relies on customized data-processing workflows that require fluorescence background removal and prior identification of Raman peaks of interest—introducing potential biases and inconsistencies. Here, we introduce DIVA (Deep-learning-based Investigation of Vibrational Raman spectra for plant-stress Analysis), a fully automated workflow based on a variational autoencoder. Unlike conventional approaches, DIVA processes native Raman spectra—including fluorescence backgrounds—without manual preprocessing, identifying and quantifying significant spectral features in an unbiased manner. We applied DIVA to detect a range of plant stresses, including abiotic (shading, high light intensity, high temperature) and biotic stressors (bacterial infections). 
By integrating deep learning with vibrational spectroscopy, DIVA paves the way for AI-driven plant health assessment, fostering more resilient and sustainable agricultural practices.
\end{abstract}

\pagebreak
Recent advances in precision agriculture---in particular, the development of data acquisition and diagnostic technologies\cite{lew2020species}---have significantly improved our ability to monitor plant health. 
A key enabler of this progress has been the development of techniques for the quantitative and non-invasive detection of important biomolecules within plants\cite{coatsworth2023continuous,sanchez2020raman,li2019non}. 
These biomolecules provide crucial information on the condition of the plants, including their exposure and subsequent response to various stressors\cite{sng2020rapid}. 
These stresses can be either abiotic\cite{payne2021raman,suzuki2014abiotic}---caused by non-living factors such as low and high light conditions, extreme temperatures, drought, waterlogging, or nutrient deficiencies---or biotic\cite{payne2021raman,suzuki2014abiotic}---caused by living organisms such as pathogens and pests. Since these stresses can significantly impact plant growth, overall health, and yield, their early detection is critical for effective crop management\cite{ang2024decoding}.

To better understand and manage plant stress, advanced sensing technologies have been proposed to enable real-time monitoring of plants in modern agricultural practices\cite{ang2024decoding,lowry2024towards,steeneken2023sensors,son2023vivo,lew2020real,lew2020species,farber2019advanced,butler2016using}.
In particular, Raman spectroscopy has been shown to be able to identify biomolecules at key metabolic centers in a plant, such as leaves and petioles\cite{chung2021rapid,sng2020rapid,butler2016using} by analyzing their unique vibrational signatures\cite{movasaghi2007raman,butler2016using}. This makes it possible to accurately assess plant health and detect early stress. 
Moreover, portable Raman spectroscopy systems\cite{weng2021advanced,gupta2020portable}, which can be easily deployed in the field, have proven particularly useful for acquiring data from hard-to-access metabolic sites within plants, providing deeper insights into their condition.

However, despite recent progress in the application of Raman spectroscopy to monitor plant health, several challenges have seriously limited its application \cite{qi2023recent}. In particular, processing Raman spectra still requires manual intervention, which can introduce biases, inconsistencies, and errors. First, custom methods are usually needed to remove background fluorescence, which would otherwise obscure the Raman signatures of biomolecules\cite{qi2023recent,fang2024recent,cadusch2013improved}. 
Additionally, Raman spectra need to be normalized, often by using their integrated intensity or the intensity of a single wavenumber as a reference, which assumes that these intensities remain stable across plant states. Finally, it is necessary to identify the specific wavenumbers corresponding to the biomolecules of interest within the Raman spectra\cite{sng2020rapid}. This process typically requires prior knowledge about which biomolecules are most likely to be affected by a particular type of stress, limiting the ability to detect broader, unexpected changes in the plant's metabolic profile. 
This narrow focus on specific biomolecules reduces the ability to capture changes throughout the spectrum, thus overlooking unexpected stress responses in the plant.
For this reason, the current approaches used to process Raman spectra are often stress-specific and designed to detect changes in a specific small set of biomolecules\cite{sng2020rapid}. 
Efforts have been made to standardize Raman data processing\cite{georgiev2024ramanspy}, such as by using fixed parameters to remove the baseline and smooth the spectrum\cite{qi2023recent,sng2020rapid,chung2021rapid}. 
Still, even these parameters can vary from user to user, introducing bias and inconsistency\cite{qi2023recent,zhao2007automated}.
These challenges reduce the overall reliability of Raman spectroscopy in detecting plant stress in a generalized way.

In order to fully realize the potential of Raman spectroscopy as a technology to detect plant stress, these limitations must be overcome.
Therefore, advances are needed in automating Raman spectral processing, reducing manual intervention, and developing more generalizable approaches to automatically detect a wide range of biomolecules under different stress conditions. 
Improving these aspects will help enhance the reproducibility, reliability, and overall utility of Raman spectroscopy, particularly within precision agriculture.

Here, we propose DIVA, a fully automated technique that uses a variational autoencoder to process native Raman spectra, including fluorescence background, to identify and quantify significant Raman peaks. 
We have successfully applied DIVA to detect abiotic stresses (due to shading, various light conditions, and high temperature) and biotic stresses (due to bacterial infections). 
By comparing the Raman spectra of plants under normal and stressed conditions, DIVA automatically identifies the significant Raman peaks, enabling unbiased identification of stress markers and ultimately paving the way for more resilient agricultural practices and smarter AI-driven plant health monitoring systems.

\section*{Results}

\subsection{DIVA to analyze Raman spectra.}

In a typical application of Raman spectroscopy to monitor plant health, Raman spectra are acquired from healthy and stressed plants, specifically from key metabolic areas such as leaves and petioles\cite{sng2020rapid}, as schematically illustrated in Fig.~\ref{Fig1}a. 
These spectra provide molecular fingerprints that reflect the presence and abundance of biomolecular signatures associated with various metabolic processes, thus indicating plant health\cite{payne2021raman,weng2021advanced}. 
In healthy plants, specific biomolecules appear within expected intensity ranges, whereas stressed plants often show distinct variations, allowing differentiation based on these unique spectral patterns.

Standard Raman spectra processing involves several preparatory steps, including baseline correction and removal of the fluorescence background noise commonly present in biological samples\cite{afseth2006raman,ryabchykov2019analyzing,qi2023recent} (Figs.~\ref{Fig1}b-c). 
This background fluorescence can overwhelm the actual Raman signals, complicating interpretation and often requiring custom correction steps\cite{cadusch2013improved,qi2023recent}.
However, fluorescence removal methods vary and can lead to inconsistent results, as different techniques may emphasize or obscure distinct parts of the spectrum.

Instead, DIVA does not require manually removing the fluorescence background and identifying a priori the relevant Raman peaks. 
It operates in two key steps.
The first step reduces the influence of fluorescence by calculating the first derivative, $D(\tilde{v})$, of the raw Raman spectra, $I(\tilde{v})$, where $\tilde{v}$ represents the wavenumbers at which the Raman intensity is acquired.
Using this derivative, we avoid the need for background correction and normalization, as the derivative of the slowly varying fluorescence background is smaller than the derivative of the fast varying Raman peaks without requiring ad hoc background fitting procedures (Fig.~\ref{Fig1}d). 
This approach enhances the spectral features, making subtle shifts in biomolecular signatures easier to detect.

The second step provides an automated analysis of these differentiated spectra using a variational autoencoder (VAE). 
The VAE’s encoder (depicted in Fig.~\ref{Fig1}e) transforms each input spectrum into a point in the latent space, condensing the complex spectral information into a lower-dimensional representation while retaining the critical characteristics of the original data (Fig.~\ref{Fig1}f). 
This latent space organizes similar spectra into clusters based on shared features, creating distinct groupings that correlate with different stressed states of the plant. 
The latent space generated by the VAE has two critical properties: continuity and completeness. 
Continuity ensures that closely placed points in the latent space yield similar decoded outputs, preserving the relationship between similar spectra. 
Completeness ensures that every sampled point within a cluster in the latent space corresponds to a meaningful spectral representation, providing a strong foundation for interpretation without losing essential data from the original Raman spectra.
From this latent space, the VAE's decoder (Fig.~\ref{Fig1}g) reconstructs the spectral content by focusing on the median of each cluster. 
This approach is important because it allows us to accurately sample representative points from each cluster, allowing the reproduction of the average characteristic spectra, $D(\tilde{v})$ (Fig.~\ref{Fig1}h), representing the clusters in the latent space.
This reconstruction captures key biomolecular signatures indicative of plant stress, providing a reliable indicator of stress conditions.

Our method for detecting significant Raman peaks for a particular stressed state of the plant relies on identifying zero crossings in the reconstructed characteristic derivative of the Raman spectrum $D(\tilde{v})$ (Fig.~\ref{Fig1}h), specifically focusing on crossings where the derivative changes from positive to negative. 
These particular crossings indicate the locations of peaks in the original spectra, I($\tilde{v}$).
We apply this technique to the reconstructed spectra, with one spectrum generated for each cluster. 
An important aspect of this approach is that it does not require any prior assumptions or hypotheses about which peaks to analyze, 
and therefore it provides a direct and systematic means of identifying peaks that correlate with important biomolecules. 
Furthermore, by calculating the area under the curve, A($\tilde{v}$), at these peak positions (illustrated in Fig.~\ref{Fig1}i), we obtain a quantitative measure of the concentrations of the biomolecules at specific wavenumbers. 
This feature of DIVA improves reproducibility and accuracy by eliminating the need for manual peak assignment, which is prone to user bias.

Finally, DIVA enables comparisons of different plant stress conditions by evaluating A($\tilde{v}$) across various states. 
Identifying and comparing the most significant Raman peaks for each state (shown in Fig.~\ref{Fig1}i), we can evaluate the intensities of these peaks at relevant wavenumbers, allowing for a comprehensive assessment of stress levels under different environmental and biological stress conditions. 
This method provides valuable insights into how plants respond to stress on a biomolecular level and establishes an unbiased framework for monitoring plant health using Raman spectroscopy.

To demonstrate the versatility of DIVA, we examined several common stresses that affect plants in traditional farming. 
We categorized these stresses into two types: abiotic (e.g., stresses due to shade avoidance, varying light conditions, and high temperature) and biotic (e.g., stress due to bacterial infection). 
Our studies included various plant species, such as Arabidopsis, Choy Sum, and Kai Lan, showcasing the broad applicability of our approach across different species. 

\subsection{Decoding light stress responses in diverse plant species.}

We began by exploring light stress as a key example of an abiotic factor to evaluate how well DIVA performs across different conditions and plant species. 
In this study, we first used Arabidopsis plants exposed to four light conditions: white light (as the control), high light, low light, and shade.
Wildtype Arabidopsis grown under these light stresses developed distinct phenotypes, reflecting how plants respond to different light stresses. 
As shown in Supplementary~Fig.~1, high light led to shorter petioles and large curled leaves, while low light and shade caused petiole elongation and reduced leaf growth. 

The latent space derived from the Arabidopsis leaf data under these conditions is shown in Fig.~\ref{Fig2}a. 
Looking at the medians of the clusters, a clear trend was observed: as the light intensity decreased from high light to white light, low light, and shade, the median points were well separated and gradually shifted from left to right. 
This shows that the molecular composition of the leaves changed systematically depending on the light levels, with brighter light conditions producing distinct molecular signatures compared to lower light conditions.

We then reconstructed the spectra from the median points of each light condition in the latent space. 
The reconstructed spectra are shown in Fig.~\ref{Fig2}b (top).
To better understand the plant response, we calculated the absolute area under the curve for each peak along the spectrum. 
This allowed us to identify the most significant peaks that characterize the response under each light condition, shown in Fig.~\ref{Fig2}b (bottom). 
These peaks were associated with key biomolecules, including carotenoids (1521 cm\textsuperscript{-1}, 1151 cm\textsuperscript{-1}, and 1180 cm\textsuperscript{-1}), cellulose/lignin/protein (1318 cm\textsuperscript{-1}), and pectin (742 cm\textsuperscript{-1})\cite{chung2021rapid}. 
The significant Raman peaks matched those seen when the Raman spectra (shown in Supplementary~Fig.~2) were reconstructed from their derivatives, reflecting consistent trends and supporting the reliability of the method.

Under different light conditions---high light, white light, low light, and shade---plants undergo coordinated changes in biomolecules such as carotenoids, cellulose, lignin, protein, and pectin to adapt their growth and mitigate stress. 
Carotenoids, in particular, were found to be strongly modulated under different light conditions. 
Carotenoids play a key role in the protection of plants by dissipating excess light energy and preventing photo-oxidative damage\cite{shumbe2014dihydroactinidiolide}. 
In high light conditions, plants increase carotenoid levels to protect against excess light stress. 
In low light and shade, the need for this protection decreases, so plants reduce carotenoid production to conserve energy and focus on improving light capture. 
In high light, cellulose and lignin levels generally increase to reinforce the cell wall and protect against photodamage, while protein levels increase, especially stress-related proteins such as antioxidants and repair enzymes.
The pectin content can also increase slightly or become structurally modified to maintain the rigidity of the cell walls and protect against oxidative damage.
In contrast, under low light and shade, cellulose and lignin levels typically decrease, allowing the plant to conserve energy on cell wall reinforcement and maintain more flexible walls to support elongation growth.
Protein levels may either decrease overall---especially with a reduction in photosynthesis-related proteins---or shift in composition toward enzymes involved in cell wall remodeling and growth regulation. 
Pectin levels also decrease or change in composition to make the cell wall more flexible, allowing faster stem and petiole elongation---as part of the shade avoidance strategy.
These molecular adjustments, summarized in Supplementary~Table~1, reflect the plant’s ability to dynamically balance structural integrity and growth depending on the surrounding light environment. 
The ability of DIVA to detect and track these changes across different light environments highlights how the method can capture biologically meaningful molecular regulation in plants---a key part of their photoacclimation strategies under varying light conditions.

Next, we extended the light stress study to include Choy Sum and Kai Lan, to test whether DIVA could capture light stress-related molecular signatures across species beyond Arabidopsis. 
The phenotypes of Choy Sum and Kai Lan under light stress are shown in Supplementary~Fig.~1. 
These two species, both leafy vegetables but with different growth habits and environmental adaptations, provided a useful comparison to assess how broadly DIVA could detect and differentiate light stress responses. 

The general pattern of median shifts and the significant Raman peaks identified by DIVA in response to light stress were consistent between Choy Sum and Kai Lan, similar to Arabidopsis as evident from Figs.~\ref{Fig2}c-d and Figs.~\ref{Fig2}e-f. 
Interestingly, the clusters for Choy Sum and Kai Lan were less compact, as observed in Figs.~\ref{Fig2}c and \ref{Fig2}e, especially under low light and shade, indicating a greater variation in their Raman spectra between individual plants compared to Arabidopsis. 
For both Choy Sum and Kai Lan, the medians for white light and high light clusters were well separated from those of low light and shade. 
The low light and shade medians were much closer to each other, suggesting that the plant response to these two light conditions was similar. 
These conditions are known to trigger petiole elongation and reduced leaf blade growth (Supplementary~Fig.~1), which is consistent with the observed clustering pattern.
Significant peaks associated with carotenoids (1521 cm\textsuperscript{-1}, 1150 cm\textsuperscript{-1}, and 1178 cm\textsuperscript{-1}), cellulose / lignin / protein (1317 cm\textsuperscript{-1}), and pectin (743 cm\textsuperscript{-1})\cite{chung2021rapid} were also observed for both Choy Sum and Kai Lan, as shown in Figs.~\ref{Fig2}d and \ref{Fig2}f.
However, the amplitude of the significant peaks varied between species, as can be observed in Figs.~\ref{Fig2}b, \ref{Fig2}d, and \ref{Fig2}f and Supplementary~Figs.~2-4, reflecting species-specific differences in molecular responses to light stress. 

In general, consistent trends across all three species highlight the potential of DIVA to track light stress responses in diverse plants. 
These findings demonstrate that DIVA can capture both species-specific responses and common molecular patterns, showcasing its robustness in identifying shared and unique adaptations to light stress. Refer to ``Light stress study" in Methods for experiment details. 

\subsection{Decoding gene-specific shade avoidance stress responses.}
To evaluate the robustness of DIVA in the capture of stress-induced molecular changes in plants, we extended our analysis to include genotypes with impaired photoreceptor function, focusing specifically on their responses to shade avoidance stress. 
Specifically, we examined whether DIVA could detect shade avoidance responses in mutant Arabidopsis lines, providing insight into how genetic variation influences stress signatures.

For this study, we first used wildtype Arabidopsis plants exposed to three light conditions: white light (control), moderate shade, and deep shade.
Wildtype Arabidopsis developed different phenotypes, reflecting how plants adjust their growth to different light conditions.
As shown in Supplementary~Fig.~5, moderate and deep shade caused elongated petioles and smaller leaf blades, typical shade avoidance responses.

The latent space derived from the Arabidopsis leaf data under these light conditions is shown in Fig.~\ref{Fig3}a.
Looking at the medians of the clusters, a clear upward shift was observed as the intensity of the shade increased from moderate to deep shade.
This indicates that the molecular composition of the leaves changed progressively under harsher shade conditions.
Raman peaks related to carotenoids (1521 cm\textsuperscript{-1}, 1151 cm\textsuperscript{-1}, and 1180 cm\textsuperscript{-1}), cellulose / lignin / protein (1318 cm\textsuperscript{-1}), and pectin (742 cm\textsuperscript{-1}) were highly responsive in different shades, as shown in Fig.~\ref{Fig3}b and  Supplementary~Fig.~6.

Under white light, Arabidopsis plants maintain balanced levels of carotenoids, cellulose, lignin, proteins, and pectin to support stable growth and photosynthesis.
In moderate shade, the photosynthetically active radiation is lower than in white light but still sufficient for photosynthesis, though at reduced intensity. 
The red-to-far-red (R:FR) light ratio decreases and plants initiate shade avoidance responses. 
Carotenoids decrease slightly because there is less need for photoprotection. 
Cellulose and lignin production decreases slightly to allow cell wall flexibility and support elongation. 
Photosynthesis-related proteins decrease, while growth-regulating enzymes become more prominent.
The pectin content decreases slightly or changes composition to enhance the flexibility of the cell wall, allowing growth of the stem and petioles.
In deep shade, the photosynthetically active radiation is much lower, often just a fraction of the radiation available in white light, making it difficult for plants to perform photosynthesis efficiently. 
The R:FR ratio drops significantly, intensifying shade avoidance responses in plants.

We then examined two light-signaling Arabidopsis mutants to investigate how different genotypes respond to shade avoidance stress and whether DIVA can capture stress-related signatures from mutant plants.
The first mutant \textit{phyA-211} knocks out Phytochrome A (PHYA), which is responsible for suppressing the shade response under low R:FR light\cite{martinez2014shade}, such as in shaded conditions. 
The second mutant \textit{phyB-9\textsuperscript{BC}} knocks out Phytochrome B (PHYB), which is responsible for suppressing the shade response under high R:FR conditions such as white light \cite{martinez2014shade}.

PHYA and PHYB play an essential role in controlling shade avoidance stress in Arabidopsis\cite{sng2020rapid}.   
PHYA helps detect low R:FR light and reduces the plant’s response to shade by preventing excessive seedling elongation. 
Without PHYA, the plant becomes more sensitive to shade, such as under moderate and deep shade conditions, especially sensitive in deep shade. 
Previous studies have shown that the \textit{phyA-211} mutant behaves like the wildtype under white light but exhibits stronger shade avoidance responses under shaded conditions\cite{sng2020rapid}. 
Consistent with this, \textit{phyA-211} plants displayed more pronounced shade phenotypes, such as elongated petioles and smaller leaves—under shade (see Supplementary~Fig.~5), confirming that the loss of PHYA leads to a more severe shade avoidance response.
This is evident in the latent space (see Fig.~\ref{Fig3}c), where the medians of the phyA-211 samples in shade shift further away from the median of the phyA-211 samples under white light.
The significant Raman peaks observed were similar to those of the wildtype, involving carotenoids, cellulose / lignin / proteins, and pectin, as shown in Fig.~\ref{Fig3}d. 
In particular, carotenoid peaks, especially at 1521 cm\textsuperscript{-1} and 1151 cm\textsuperscript{-1}, were significantly impacted in the \textit{phyA-211} mutant under deep shade, likely due to enhanced light stress responses to deep shade in the absence of PHYA (see Fig.~\ref{Fig3}d and Supplementary~Fig.~7).

PHYB is active in high R:FR light and helps suppress shade avoidance traits such as petiole elongation and reduced leaf blade size. 
When PHYB is knocked out, as in the \textit{phyB-9\textsuperscript{BC}} mutant, this suppression is lost, leading plants to show a strong shade avoidance behavior even under white light. 
This is reflected in the typical shade phenotypes of the mutant, such as elongated petioles and small rosette leaves, observed under white light conditions (see Supplementary~Fig.~5).
This pattern is also reflected in the latent space (see Fig.~\ref{Fig3}e), where the medians of \textit{phyB-9\textsuperscript{BC}} plants in white light and deep shade are closely located. 
The significant Raman peaks observed for \textit{phyB-9\textsuperscript{BC}} plants were comparable to those of the wildtype, involving carotenoids, cellulose / lignin / proteins and pectin. 
However, in both white light and deep shade, the biomolecular profiles remained similar, as seen in Fig.~\ref{Fig3}f and Supplementary~Fig.~8, due to the constitutive shade response of \textit{phyB-9\textsuperscript{BC}} plants. 
Notably, the medians for all three conditions tended to shift toward the upper left corner of the latent space, highlighting that the PHYB variant behaves in a similar way across different light conditions.
Furthermore, based on the location of the medians, the responses of \textit{phyB-9\textsuperscript{BC}} plants to white light and deep shade were similar to each other, but distinct from their moderate shade response (see Fig.~\ref{Fig3}f and Supplementary~Fig.~8). 
This suggests that while PHYB plays an important role in regulating shade responses, it is not acting alone---other signaling components or pathways likely contribute to how plants respond to varying light environments.


Overall, the findings matched the expected trends in both wildtype and mutant plants under shade avoidance stress, underscoring DIVA’s ability to monitor responses across diverse genotypes. 
DIVA reliably identified genotype-specific differences and common molecular patterns, demonstrating its strength in revealing shared and distinct adaptations to shade stress.
Refer to ``Shade avoidance stress study" in Methods for experiment details.

\subsection{Decoding high-temperature stress responses.}

Understanding how temperature extremes affect plant physiology is critical to improving crop resilience. 
To evaluate whether DIVA can detect species-specific molecular responses to heat, we studied wildtype Arabidopsis, Choy Sum, and Kai Lan plants exposed to high ambient temperatures. 
This cross-species analysis aimed to uncover how different plants adapt at the molecular level to prolonged heat stress.

Wildtype Arabidopsis plants were initially grown at 21\textsuperscript{o}C (control) and then exposed to 29\textsuperscript{o}C for varying durations, as outlined in Supplementary~Fig.~9. 
All plants were in the same developmental stage at the end of treatment. 
As shown in Supplementary~Fig.~10, heat-exposed Arabidopsis plants developed elongated petioles---a classic thermomorphogenic response that increases leaf separation to improve air circulation and reduce localized heat build-up.

The latent space derived from Arabidopsis leaf data under control and high-temperature conditions is shown in Fig.~\ref{Fig4}a. 
After 2 days of high-temperature treatment, the median position shifted away from the control, indicating early biomolecular changes. 
By day 4, the median remained close to the day-2 position, suggesting a limited additional change. 
By day 6, the data split into two distinct clusters---one at the top and one at the bottom---reflecting varied responses, with only some plants showing visible signs of yellowing and senescence.
This divergence likely reflects differing physiological states: while some plants maintained biomolecular balance under sustained heat, others crossed a physiological threshold leading to structural and metabolic decline.
 
These shifts in the latent space implied clear changes in key biomolecules.
Raman peaks, especially for carotenoids at 1521 cm\textsuperscript{-1}, 1151 cm\textsuperscript{-1}, and 1180 cm\textsuperscript{-1}, cellulose / lignin / protein at 1318 cm\textsuperscript{-1}, and pectin at 742 cm\textsuperscript{-1}, were markedly responsive, as evident from Fig.~\ref{Fig4}b and Supplementary~Fig.~11.
Carotenoids decreased, probably due to oxidative damage and reduced protective capacity against reactive oxygen species. 
Structural components such as cellulose, lignin, and proteins also declined, reflecting heat-induced degradation and possible remodeling of the cell wall. 
Pectin levels showed a similar drop, suggesting a loosening of cell walls as part of the adaptive strategy of plants. 
However, prolonged or intense heat can lead to excessive degradation in Arabidopsis, compromising structural integrity and contributing to wilting and/or senescence.
This expectation aligned with the observations on day 6, where the latent space data split into two distinct clusters---one group representing plants that maintained molecular stability, and another showing clear signs of senescence, including yellowing (see Fig.~\ref{Fig4}a). 
To illustrate these differences, the representative Raman peaks shown for the day 6 time point in Fig.~\ref{Fig4}b reflect the overall median signature of the entire group of the day 6 time point (see Fig.~\ref{Fig4}a), without excluding any data points.
This highlights DIVA’s strength in uncovering subtle but meaningful intragroup variations.
Together, these observations demonstrate that while Arabidopsis can mount an early response to high temperature, prolonged exposure eventually impairs its ability to maintain essential molecular functions.

In contrast, Choy Sum plants appeared to be more tolerant to high temperatures.
Choy Sum plants were grown at 24\textsuperscript{o}C and then exposed to 33\textsuperscript{o}C for varying durations, as shown in Supplementary~Fig.~12. 
All plants were in the same developmental stage at the end of treatment. 
Phenotypic observations showed that Choy Sum exhibited minimal signs of heat-induced stress, with little visible thermomorphogenesis compared to Arabidopsis, as evident from Supplementary~Fig.~13. 

The latent space derived from Choy Sum leaf data under control and high temperature conditions is shown in Fig.~\ref{Fig4}c. 
The medians of the clusters---each representing a different time point under high-temperature treatment---were mostly positioned near the top of the latent space, with the control group forming a distinct cluster close to them.
Notably, these clusters occupied a different region of the latent space compared to Arabidopsis, highlighting the biological distinction of Choy Sum plants from Arabidopsis and the species-specific differences in their response to high temperature.
The medians of 2 to 6 day time points overlapped substantially, suggesting that the biomolecular profile of Choy Sum plants remained largely stable throughout the period of prolonged heat exposure.
This minimal spectral shift indicates a muted stress response, consistent with the phenotypic evidence of Choy Sum’s apparent tolerance to high temperatures.
In particular, while a clear separation was observed between the control and day 2 medians, the medians of subsequent time points (day 4 and day 6) remained close to the day 2 median, further supporting the idea that most of the biomolecular adjustments occurred early in the heat treatment period.

In Choy Sum, the significant Raman peaks identified under high temperature treatment are shown in Fig.~\ref{Fig4}d. They were consistent with those observed earlier: carotenoids at 1521 cm\textsuperscript{-1}, 1150 cm\textsuperscript{-1}, and 1178 cm\textsuperscript{-1}; cellulose/lignin/protein at 1317 cm\textsuperscript{-1}; and pectin at 743 cm\textsuperscript{-1}. 
These peaks showed a marked increase in intensity from control to day 2, followed by only minimal changes by days 4 and 6. 
This pattern indicates an early biomolecular adjustment to high temperature, with limited further modulation over time. 
The same trend was evident in the reconstructed Raman spectra (see Supplementary~Fig.~14) obtained from the cluster medians in the latent space, further confirming the observation of early but sustained biomolecular responses.
This early upward shift in biomolecular levels probably reflects the enhanced accumulation or preservation of these compounds in response to high temperatures. 
Carotenoids may have increased due to their role in photoprotection and quenching of reactive oxygen species, buffering the plant against oxidative stress. 
Similarly, the rise in cellulose, lignin, and protein-associated signals points to the reinforcement of the cell walls, contributing to mechanical resilience.
The elevated pectin signals suggest ongoing cell wall remodeling to maintain cellular integrity.

Taken together, the location of the cluster medians in the latent space, the early biomolecular increase, and the sustained Raman intensity levels over time suggest that Choy Sum employs a front-loaded molecular strategy to cope with heat---markedly different from the continuous decline observed in Arabidopsis. 

Lastly, we examined Kai Lan plants to further understand species-specific responses to high temperature.
Kai Lan was grown under the same conditions as Choy Sum (see Supplementary~Fig.~12).
Kai Lan exhibited a distinct physiological and biomolecular response to heat stress, with phenotypic signs such as yellowing of mature leaves and purple coloration in young ones, as shown in Supplementary~Fig.~15.

The latent space derived from Kai Lan leaf data under control and high-temperature conditions is shown in Fig.~\ref{Fig4}e. 
The clusters representing heat-treated time points (days 2, 4, and 6) were distributed along the lower and right regions of the latent space, contrasting with the locations of these clusters observed in Arabidopsis (see Fig.~\ref{Fig4}a) and Choy Sum (see Fig.~\ref{Fig4}c).
Notably, the progressive shift in cluster medians---from minimal displacement at day 2 to clear separation at day 6---indicates a cumulative and dynamic biomolecular response to heat stress.

This trajectory was mirrored in the significant Raman peaks shown in Fig.~\ref{Fig4}f, where the related biomolecular content showed a gradual increase over time. 
These peaks, consistent with those seen in Choy Sum, include carotenoids at 1521 cm\textsuperscript{-1}, 1150 cm\textsuperscript{-1}, and 1178 cm\textsuperscript{-1}; cellulose / lignin / protein at 1317 cm\textsuperscript{-1}; and pectin at 743 cm\textsuperscript{-1}. 
Unlike Choy Sum, where most biomolecular changes occurred early, Kai Lan showed a stepwise increase in Raman peak intensities from day 2 to day 6, indicating a continued and evolving response to prolonged heat exposure. 
This temporal progression was also clearly reflected in the reconstructed Raman spectra (see Supplementary~Fig.~16) obtained from the latent space medians, supporting the DIVA-based observations.

Together, the progressive shifts of the medians in the latent space, the steadily increasing significant Raman peaks, and the Raman intensities in the reconstructed spectra indicate that Kai Lan responds to heat stress through a cumulative biomolecular adaptation. 
This contrasts with the early stabilized response seen in Choy Sum and the overall decline observed in Arabidopsis. 
These distinctions underscore Kai Lan’s unique and dynamic strategy to cope with prolonged high temperatures and highlight DIVA's sensitivity in capturing nuanced species-specific stress trajectories.
Refer to ``High-temperature stress study" in Methods for experiment details.

\subsection{Decoding bacterial-infection induced stress responses.}

Understanding how plants respond to biotic stress, such as bacterial infection, is critical to improving crop protection and resistance. 
To explore how plants react at the molecular level to bacterial pathogens and immune elicitors, we applied DIVA to study Choy Sum and Arabidopsis plants infiltrated with different bacterial treatments over time.

To investigate pathogen-induced responses, Choy Sum plants were infiltrated with a bacterial suspension, while buffer-infiltrated plants served as controls at each time point. 
Raman spectra were collected from leaves 24 and 48 hours post-infiltration to monitor biomolecular changes during the course of infection. 
The corresponding leaf phenotypes are shown in Supplementary~Fig.~17. 
The buffer- and pathogen-infiltrated leaves appeared healthy, with no visible signs of infection observable 48 hours post-infiltration.

The latent space derived from the 24-hour post-infiltration data is shown in Fig.~\ref{Fig5}a. 
The medians of the pathogen- and buffer-infiltrated groups were slightly separated, indicating early molecular divergence. 
The extracted features in Fig.~\ref{Fig5}b and the reconstructed Raman signals in Supplementary~Fig.~18 revealed specific biomolecular changes, with significant peaks detected for carotenoids (1521 cm\textsuperscript{-1} and 1150 cm\textsuperscript{-1}), cellulose / lignin / proteins (1317 cm\textsuperscript{-1}), and pectin (743 cm\textsuperscript{-1}). 

By 24 hours post-infiltration, carotenoid levels had decreased slightly in pathogen-infiltrated plants compared to buffer-infiltrated controls. 
This modest reduction may have reflected the early mobilization of carotenoids to counteract the oxidative stress induced by the presence of pathogens. 
In contrast, compounds such as cellulose, lignin, protein, and pectin exhibited minimal changes, possibly indicating that structural remodeling or large-scale metabolic changes had not yet begun.
Interestingly, a Raman peak at 1001 cm\textsuperscript{-1} was significant only in buffer-infiltrated plants, while a peak at 1178 cm\textsuperscript{-1} appeared significant only in pathogen-infiltrated plants. 
This suggested the presence of a stress-induced metabolite or a marker of a specific bacterial response.
A plausible explanation for this pattern is that 24 hours post-infiltration, the 1001 cm\textsuperscript{-1} peak was significant in buffer-infiltrated plants due to a transient, nonspecific stress response caused by mechanical infiltration. 
This peak, associated with phenylalanine and carotenoids, probably reflected early warning signals or mild oxidative stress in the absence of pathogens.
In pathogen-infiltrated plants, the 1178 cm\textsuperscript{-1} peak was significant, likely indicating an early pathogen-specific activation of carotenoid metabolism in response to bacterial-induced oxidative stress. 
Although the 1001 cm\textsuperscript{-1} peak---representing general stress signals such as phenylalanine---may also have been present in the pathogen group, the stronger carotenoid-related response at 1178 cm\textsuperscript{-1} probably dominated the Raman signal. 
This suggests that while both groups experienced early stress, bacterial infection triggered a more specific and pronounced response that may have masked or shifted general stress signals.

By 48 hours, the separation between the medians of the two groups increased (see Fig.~\ref{Fig5}c), indicating a stronger and more consistent biomolecular impact from bacterial infection, as evidenced by the reconstructed Raman signals in Supplementary~Fig.~19. 
However, individual data points were more spread across both groups, especially in pathogen-infiltrated samples, reflecting heterogeneous responses across different regions of the leaves. 
The significant Raman peaks observed at 48 hours were the same as those at 24 hours (see Fig.~\ref{Fig5}d). 

The decrease in carotenoid content was more pronounced in pathogen-infiltrated plants by 48 hours, suggesting a sustained or intensified response to oxidative stress. 
This continued decline was likely due to the degradation or conversion of carotenoids into signaling molecules or antioxidants as part of plant defense mechanisms. 
In contrast, relatively stable levels of cellulose, lignin, protein, and pectin indicated that the infection had not yet progressed to a stage requiring significant structural or compositional changes in the cell wall or protein content.
At this time point, the 1001 cm\textsuperscript{-1} peak became significant in pathogen-infiltrated plants, while the 1178 cm\textsuperscript{-1} peak became significant in buffer-infiltrated plants. 
The 1001 cm\textsuperscript{-1} peak probably reflected a stronger defense response involving the accumulation of protective compounds such as phenylalanine. 
The 1178 cm\textsuperscript{-1} peak in buffer-infiltrated plants may have indicated that these plants had largely recovered from the initial infiltration stress.
In general, these results suggest that buffer-infiltrated plants experienced a brief and general stress response, whereas pathogen-infected plants mounted a persistent defense reaction over time.

Taken together, these results demonstrate that DIVA can sensitively track the onset and progression of pathogen-induced stress at the molecular level---even when no visible signs of infection are present. 
The clear separation in latent space and the appearance of specific Raman peaks underscore the dynamic and time-sensitive response of the plants to the pathogen invasion.
Refer to ``Bacterial stress study" in Methods for experiment details on pathogen-induced infection in Choy Sum plants.

To explore immune-specific responses without direct pathogenic symptoms, we treated Arabidopsis plants with bacterial elicitors---specifically \textit{\textit{elf18}} and \textit{\textit{flg22}} peptides---known to induce pattern-triggered immunity\cite{chung2021rapid}. 
The plants were infiltrated with mock and bacterial elicitor suspensions and then sampled at three time points: 24, 48, and 72 hours post-infiltration. 
The mock-infiltrated plants served as controls.

At 24 hours, the latent space showed that \textit{elf18}-infiltrated plants had a well-separated median from the \textit{flg22} and mock groups (see Extended~Data~Fig.~\ref{EDFig1}a), suggesting an early and distinct immune response. 
Significant changes in carotenoids (1521 cm\textsuperscript{-1}, 1151 cm\textsuperscript{-1}, 1180 cm\textsuperscript{-1}), cellulose / lignin / protein (1318 cm\textsuperscript{-1}), and pectin (742 cm\textsuperscript{-1}) were observed in all groups, as evident from Extended~Data~Fig.~\ref{EDFig1}b and Supplementary~Fig.~20. 
These biomolecules are commonly involved in cellular signaling, structural reinforcement, and oxidative protection.
The \textit{elf18} group registered an increase in the intensity of these Raman peaks compared to the mock and \textit{flg22} groups. 
This pattern probably indicated that \textit{elf18} triggered a stronger and earlier defense response, marked by biochemical and structural changes, compared to \textit{flg22} and mock treatments.

By 48 hours, the three groups showed clearly separated medians in the latent space (see Extended~Data~Fig.~\ref{EDFig1}c), indicating that both elicitors had triggered strong molecular responses. 
The significant Raman peaks observed at 48 hours were the same as those at 24 hours (see Extended~Data~Fig.~\ref{EDFig1}d), except for a peak at 843 cm\textsuperscript{-1} specific to the \textit{elf18} group.
Notably, the \textit{flg22}-infiltrated group exhibited stronger Raman signals than the control and \textit{elf18} groups (see Extended~Data~Fig.~\ref{EDFig1}d and Supplementary~Fig.~21), suggesting a pronounced immune activation by the 48-hour time point. 
Interestingly, the \textit{elf18} group showed weaker Raman signals than the control at this stage, hinting at a possible oscillating immune behavior or molecular suppression following initial activation. 
The emergence of the 843 cm\textsuperscript{-1} peak in \textit{elf18}-treated plants may have reflected a specific biochemical response triggered by the elicitor. 
Since this peak has been associated with phospholipids---key components of cell membranes and important signaling molecules during stress---it may have indicated that immune activation had occurred at the membrane level. 
The \textit{elf18} elicitor could have initiated processes such as lipid remodeling, membrane signaling, or vesicle trafficking. 
These changes might have represented the preparation of the plant to defend against the \textit{elf18} elicitor, making the 843 cm\textsuperscript{-1} peak a potential marker of \textit{elf18}-induced immune responses at this time point.

At 72 hours, the medians of the three groups remained distinct in the latent space (see Extended~Data~Fig.~\ref{EDFig1}e). 
The significant Raman peaks observed at 72 hours were the same as those at 24 hours (see Extended~Data~Fig.~\ref{EDFig1}f and Supplementary~Fig.~22).
However, the \textit{elf18} group now showed Raman intensities stronger than the mock, suggesting a continued oscillating immune behavior and a resurgence in the immune response by the 72-hour time point. 
The \textit{flg22} group showed weaker Raman intensities than the mock, also indicating an oscillating immune behavior similar to the \textit{elf18} group. 
The pattern observed across all time points suggests that both \textit{elf18} and \textit{flg22} groups exhibit oscillating phased responses to their respective bacterial elicitors; however, \textit{elf18} triggers an earlier response, in contrast to the delayed activation seen in the \textit{flg22} group.

These findings underscore DIVA’s capability to detect elicitor-specific responses in plants, offering a powerful approach to monitor immune dynamics. 
The time-dependent shifts in latent space and Raman signatures highlight the fine-tuned nature of plant immune responses and how they vary between elicitors.
Refer to ``Bacterial stress study" in Methods for experiment details on elicitor-induced infection in Arabidopsis plants.

\section*{Discussion}

In an era where climate resilience is essential for crop development, it is critical to decode plant stress responses with precision and generalizability. 
This work introduces DIVA (Fig.~\ref{Fig1}) as an unsupervised method for analyzing Raman spectral data to reveal stress-related physiological changes in plants.

DIVA provides a unified framework for the decoding of plant stress and is built on an encoder-decoder architecture trained on the first derivatives D($\tilde{v}$) of Raman spectra. 
This design allows the model to learn from subtle spectral changes and form low-dimensional latent spaces that reflect the biomolecular state of the plants. 
By bypassing the need for predefined labels or hand-crafted features, DIVA directly interprets high-dimensional Raman signals and captures stress progression and adaptation across plant species.

DIVA demonstrated a robust capacity to capture early and nuanced molecular responses to stress before visible phenotypic changes emerged. 
For example, in the high-temperature stress study, the model revealed species-specific trajectories: Arabidopsis showed a clear decline in molecular stability, Choy Sum exhibited a stabilization trend, and Kai Lan displayed a phased adaptation over time. 
Importantly, DIVA also captured intragroup variation, most notably in Arabidopsis, highlighting divergent physiological responses between individuals subjected to the same treatment. 
These results underscore how DIVA’s latent space effectively encodes complex spectral descriptors, including those not explicitly known to the researcher, and reconstructs them with high fidelity. 
Furthermore, the ability of the model to automatically identify key stress-related peaks from the first derivative spectra adds to its interpretability and practical value for plant stress phenotyping.

DIVA's latent space can be considered a reflection of the physiology of the plant species considered, as it revealed distinct clustering behaviors that closely mirrored physiological stress responses across different species. 
For example, in Arabidopsis subjected to high temperature stress, two well-separated latent clusters emerged by day 6, corresponding to the onset of visible senescence in a subset of plants. 
In contrast, Choy Sum exhibited significant cluster overlap, indicating a degree of molecular stability under stress. 
Meanwhile, Kai Lan showed a more continuous shift in latent space, reflecting a gradual and adaptive physiological response.
These emergent spatial patterns illustrate how DIVA can trace stress trajectories and infer physiological states using the inherent structure of the data, without relying on predefined labels or supervision.

DIVA’s approach enables interpretable spectral analysis.
Training on the first derivative D($\tilde{v}$) spectra enhances the model’s sensitivity to spectral gradients, which is essential for detecting subtle biomolecular shifts.
In addition, DIVA uses area-under-the-graph data A($\tilde{v}$), derived from D($\tilde{v}$), to represent biomolecular content. 
This provides a robust and interpretable readout, as the integrated area reflects both the presence and intensity of Raman-active features.
By linking specific peaks to underlying biochemical changes, this method allows for precise identification of biomolecular responses, distinguishing DIVA from traditional Raman analysis techniques.

DIVA conserves biomarkers and species-specific strategies to handle stress. 
Despite differences in response dynamics, a core set of stress-sensitive signatures---carotenoids, proteins, and cell wall compounds---is observed. 
For example, in the study of high temperature stress, Arabidopsis experienced a steady decline, suggesting molecular breakdown; Choy Sum showed an early increase that later stabilized; and Kai Lan exhibited a gradual rise, indicating ongoing adaptation.
These patterns suggest that while biomarkers are conserved, their modulation strategies differ across species---a finding made accessible through DIVA’s ability to encode and decode spectral features under varying conditions.
In addition, DIVA demonstrated generalizability across multiple stressors. 
It generalized well across plant types and stress conditions, indicating its potential to detect and differentiate multiple stress factors within a single model. 
This flexibility makes DIVA suitable for deployment in complex real-world agricultural datasets.

Conceptually, DIVA offers several advantages over traditional dimensionality reduction and visualization methods commonly used in biological studies, such as Principal Component Analysis (PCA) and t-distributed Stochastic Neighbor Embedding (t-SNE), which lack reconstructive feedback and probabilistic modeling.
In contrast, DIVA employs unbiased learning without the need for labeled data.
Its smooth latent space captures continuous biological variation over time, while dimensionality reduction helps suppress noise and highlight key features.
Importantly, the model’s reconstructive feedback ensures that the learned latent representations remain grounded in biologically meaningful patterns.
Together, these design elements position DIVA as a promising alternative to traditional approaches for generating interpretable insights from complex spectral data.

Although DIVA offers significant advantages in unbiased spectral analysis, several limitations must be acknowledged. 
First, DIVA is inherently restricted to the detection of Raman-active compounds, which means that it currently cannot capture physiological changes associated with non-Raman-active markers, such as hormones. 
To address this, future work could involve integrating Raman spectroscopy with complementary modalities such as transcriptomics, hyperspectral imaging, or mass spectrometry, enabling a multimodal framework that captures a broader range of stress responses. 
Second, although environmental conditions such as light, humidity, and temperature were tightly controlled in our experiments, DIVA does not explicitly model these external factors.
This limitation is common across all analytical methods and could affect robustness when applied in field settings where such variables fluctuate.
Incorporating domain adaptation techniques or adversarial learning could help the model learn domain-invariant representations, enhancing its generalization to diverse and uncontrolled environments.
Finally, while DIVA’s latent space captures meaningful patterns related to the plant’s response to stress, the individual dimensions of this space are not yet directly linked to specific biological variables.
In other words, the structure of the latent space is currently interpretable in a relative or qualitative sense, rather than being quantitatively aligned with known biological stressors.
Future work could explore model extensions that aim to separate different types of variation---such as stress type, intensity, or time point---into distinct parts of the latent space.
This may help make the model’s internal representations more interpretable and biologically grounded, moving beyond abstract patterns to features that align with real-world plant responses.

This work opens several promising avenues for future research and application. 
One key direction is field deployment---by pairing DIVA with portable Raman spectrometers, it could enable real-time stress monitoring in open farms and controlled environment agriculture, making it a practical tool for precision agriculture. 
Beyond static assessment, DIVA could also be extended to predictive modeling to predict stress trajectories, allowing early intervention before irreversible damage occurs. 
Additionally, the adaptability of the DIVA framework opens the door to cross-species transfer; using transfer learning, DIVA could be fine-tuned for new crops with minimal training data, thereby broadening its applicability across agricultural systems. 
Integration into digital twinning frameworks is another compelling direction, where DIVA could contribute to real-time automated plant health diagnostics in farming systems. 
Notably, the encoder-decoder architecture of DIVA also allows the identification of previously unrecognized biomolecular patterns linked to stress, potentially uncovering novel biological insights that go beyond existing knowledge.

The broader impact of this work spans multiple disciplines. 
For plant scientists, DIVA provides a valuable analytical tool for studying latent physiological changes before the onset of visible symptoms, allowing earlier detection and a deeper understanding of stress responses. 
For AI researchers, it serves as a proof-of-concept for the application of unbiased generative models to uncover meaningful and interpretable biological insights from complex spectral data. 
From an agricultural perspective, DIVA represents a scalable, non-intrusive, and cost-effective solution for early stress detection, supporting more sustainable and precise crop management practices.

DIVA marks a step forward in the decoding of plant stress responses and in the monitoring of plant health in general. 
By capturing and reconstructing Raman spectral features in an unsupervised and interpretable way, DIVA offers a robust, generalizable method for early and accurate stress detection across plant species and conditions.
DIVA's flexibility, automation, and interpretability make it suitable for both scientific research and real-world agricultural deployment. 
DIVA provides not only a tool, but also a framework for well-informed, data-driven crop resilience strategies.

\begin{figure}
    \centerline{\includegraphics[width=\textwidth]{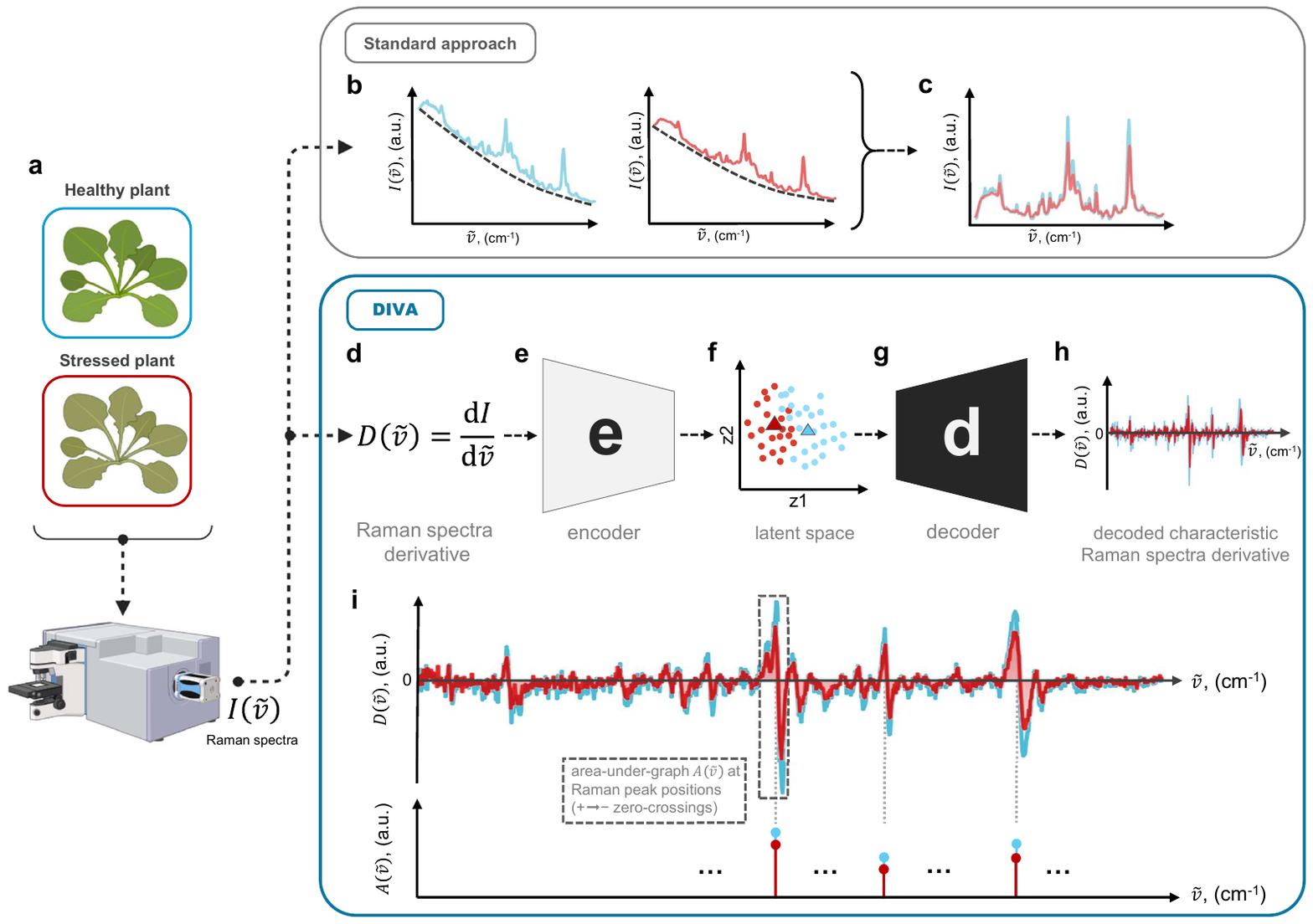}}
    \linenumbers
    \caption{\bodyfigurelabel{Fig1} 
    \textbf{Deep learning-based Investigation of Vibrational Raman spectra for plant-stress Analysis (DIVA). }
    \textbf{a}, Raw Raman spectra, I($\tilde{v}$),  collected by a Raman spectrometer, capture biomolecular fingerprints of plants at specific times. 
    \textbf{b-c}, Standard Raman signal processing requires baseline estimation (\textbf{b}) and removal from the raw spectra (\textbf{c}). 
    \textbf{d-h}, In the proposed model-free approach, the raw  Raman spectra are differentiated (\textbf{d}) and fed to a variational autoencoder (VAE). 
    The VAE encoder (\textbf{e}) is trained to map the differentiated spectra into points in a low-dimensional latent space (\textbf{f}). 
    For each plant condition (e.g., healthy or stressed), the VAE decoder (\textbf{g}) can reconstruct the average latent space of the characteristic differentiated spectra, D($\tilde{v}$) (\textbf{h}). 
    Differences between the average characteristic differentiated spectra reveal the spectral traits characteristic of each plant condition (e.g., healthy vs. stressed states). 
    \textbf{i}, The zero-crossings in the decoded characteristic differentiated spectra identify the Raman peaks whose significance can be inferred from the area under the graph, A($\tilde{v}$).
    }
\end{figure}

\begin{figure}
    \centerline{\includegraphics[width=\textwidth]{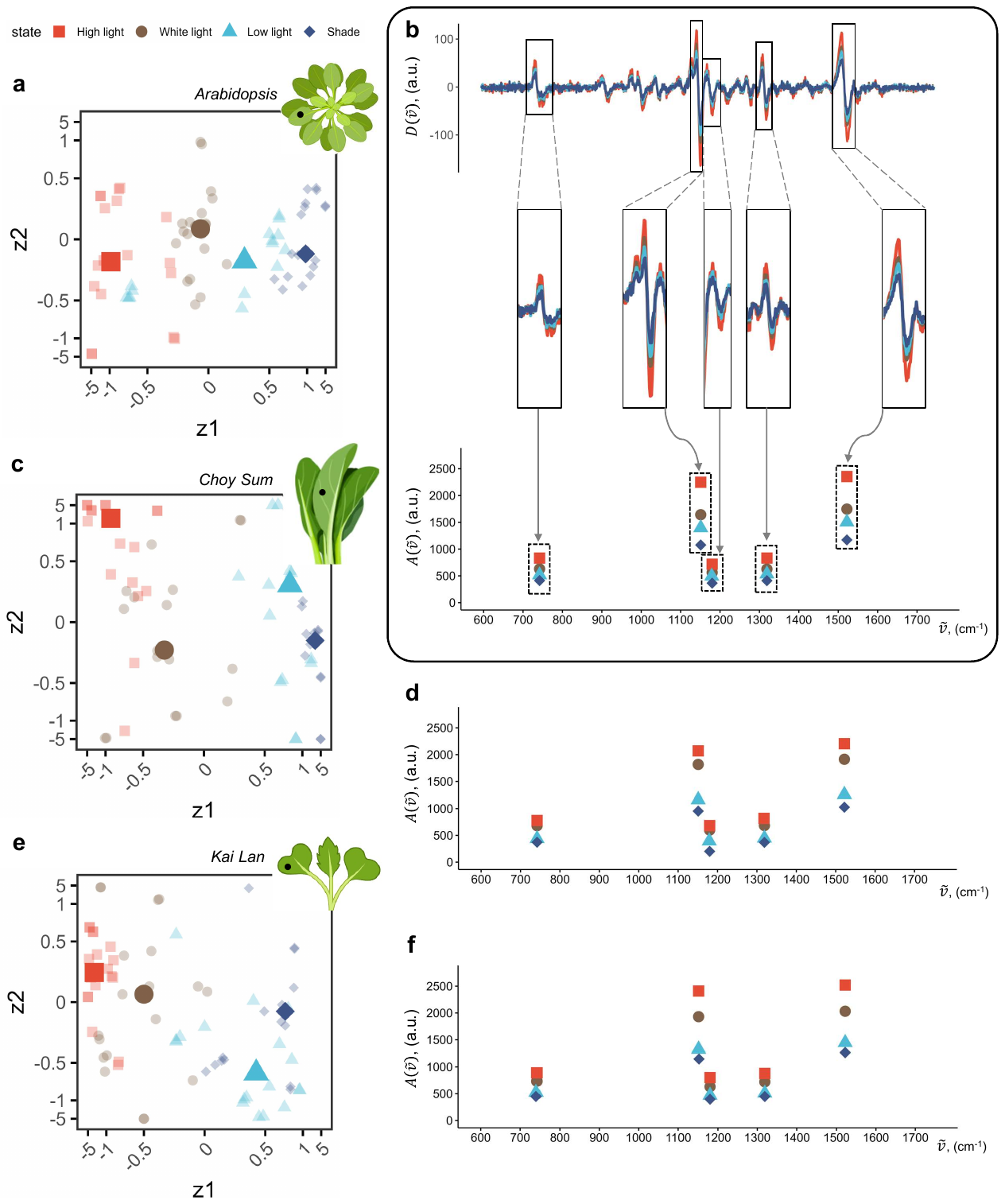}}
    \linenumbers
	\caption{\bodyfigurelabel{Fig2}
		\textbf{Decoding light stress responses in Arabidopsis, Choy Sum, and Kai Lan plants.}  Application of DIVA for light stress analysis in leaves of wildtype Arabidopsis, Choy Sum, and Kai Lan plants.
        Refer to ``Light Stress Study" in Methods for details on the light stress experiment. 
		\textbf{a}, The Raman spectra acquired from the leaves of wildtype Arabidopsis plants under white light, high light, low light, and shade conditions are passed through VAE to analyze the plant response to light stress at the spectral level. 
        The two-dimensional latent space represents the spectra acquired from the leaves of wildtype Arabidopsis plants. The white light, high light, low light, and shade clusters learned by the VAE and the cluster medians are identified.  
		\textbf{b}, Visualization of the Raman spectra derivative D($\tilde{v}$) (top) reconstructed from the medians of the clusters. The five most significant peaks for each state are also shown (bottom). 
		\textbf{c}, \textbf{a} applied to wildtype Choy Sum plants. 
        \textbf{d}, Visualization of the five most significant peaks for each state of Choy Sum plants. 
		\textbf{e-f}, \textbf{c-d} applied to wildtype Kai Lan plants.
        \textbf{b, d, f}, For all plants, the five most significant peaks identified by DIVA are carotenoids (1521 cm\textsuperscript{-1}, 1150 cm\textsuperscript{-1}, and 1180 cm\textsuperscript{-1}), cellulose, lignin, and proteins (1318 cm\textsuperscript{-1}), and pectin (742 cm\textsuperscript{-1}).
	}
\end{figure}

\begin{figure}
    \centerline{\includegraphics[width=\textwidth]{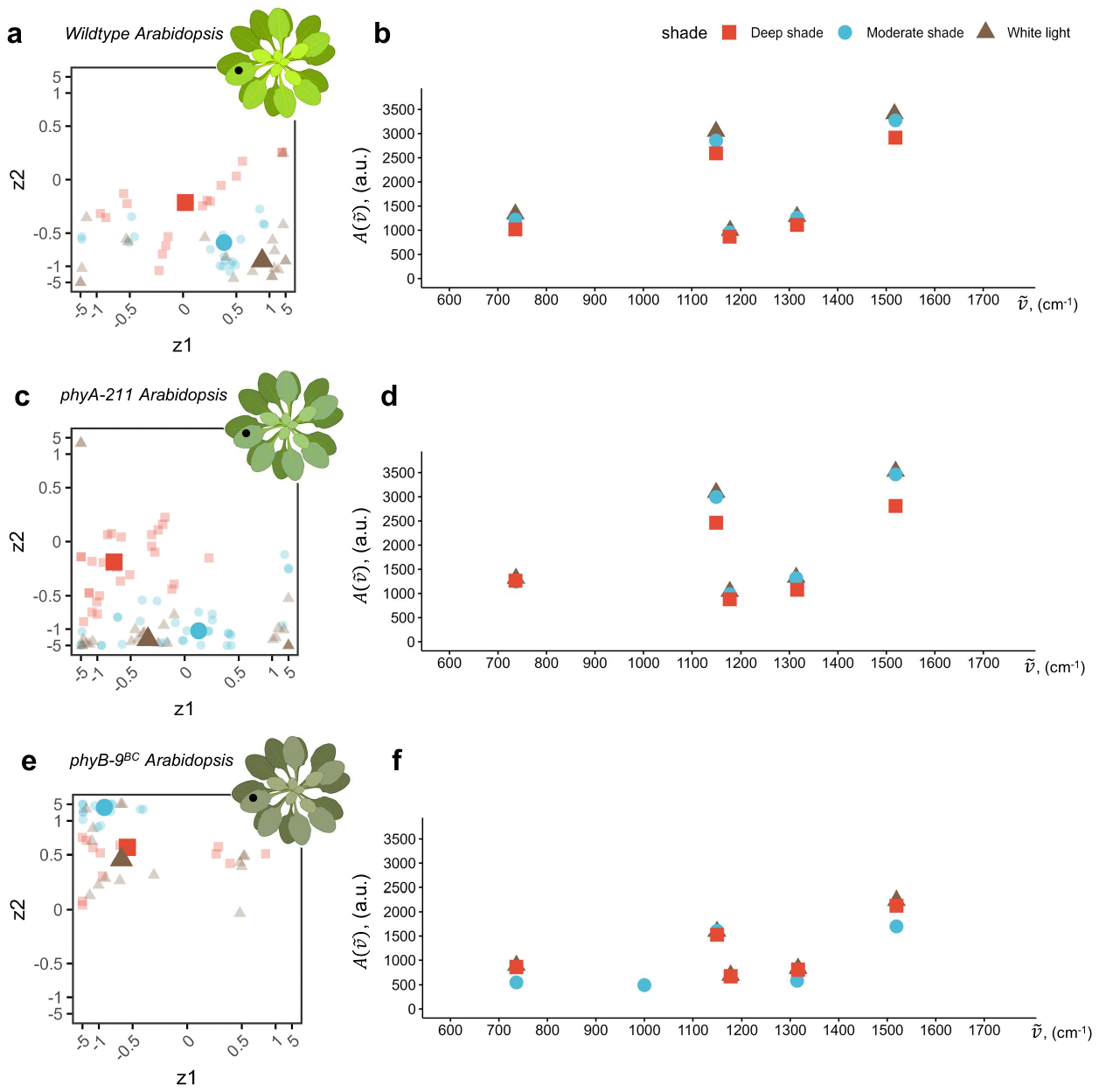}}
    \linenumbers
	\caption{\bodyfigurelabel{Fig3}
        \textbf{Decoding shade avoidance stress responses in wildtype and mutant Arabidopsis plants.}  Application of DIVA for shade avoidance stress analysis in leaves of wildtype and mutant Arabidopsis plants. 
        Refer to ``Shade avoidance stress study" in Methods for details on the shade avoidance stress experiment. 
		\textbf{a}, The Raman spectra acquired from the leaves of wildtype Arabidopsis plants under white light, moderate shade, and deep shade conditions are passed through a VAE to analyze the plant response to shade avoidance stress at the spectral level. 
        The two-dimensional latent space represents the spectra acquired from the leaves of wildtype Arabidopsis plants. The white light, moderate shade, and deep shade clusters learned by the VAE and the cluster medians are identified.  
		\textbf{b}, The five most significant peaks for each state are identified.
		\textbf{c-f}, DIVA applied to \textit{phyA-211} Arabidopsis mutant plants (\textbf{c}, \textbf{d}) and \textit{phyB-9\textsuperscript{BC}} Arabidopsis mutant plants (\textbf{e}, \textbf{f}).
	}
\end{figure}

\begin{figure}
    \centerline{\includegraphics[width=\textwidth]{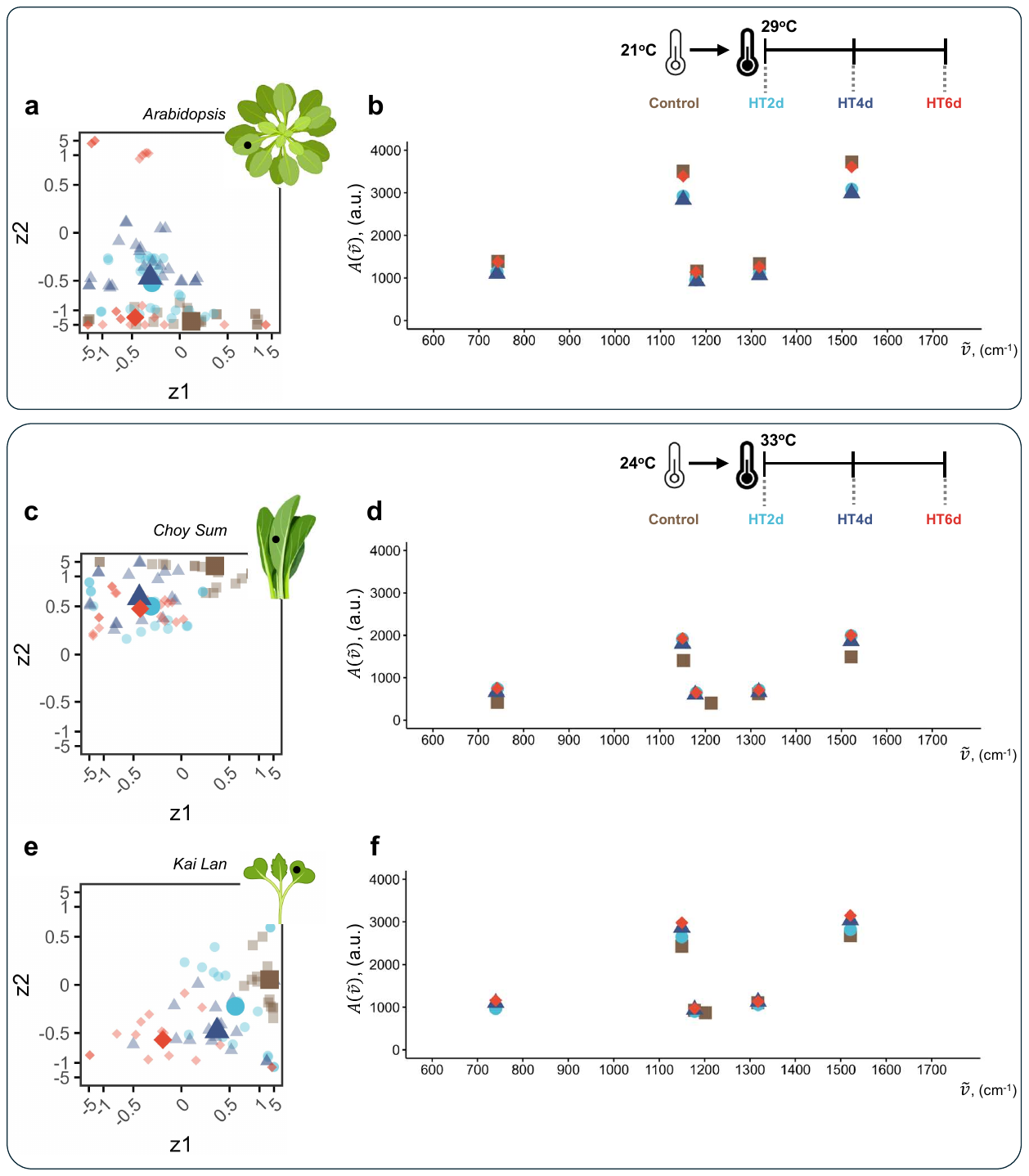}}
    \linenumbers
	\caption{\bodyfigurelabel{Fig4}
		\textbf{Decoding high-temperature stress responses in wildtype Arabidopsis, Choy Sum, and Kai Lan plants.} Application of DIVA for high-temperature stress analysis in leaves of wildtype Arabidopsis, Choy Sum, and Kai Lan plants. Refer to ``High-temperature stress study" in Methods for details on the high-temperature stress experiment. 
		\textbf{a}, The Raman spectra acquired from the leaves of wildtype Arabidopsis plants at ambient temperature (21 \textsuperscript{o}C; Control) and high-temperature (29 \textsuperscript{o}C) conditions at 2-day, 4-day, and 6-day time points (referred to as HT2d, HT4d, and HT6d, respectively) are passed through a VAE to analyze the plant response to high-temperature stress at the spectral level. 
		The two-dimensional latent space represents the spectra acquired from the leaves of wildtype Arabidopsis plants. 
        The Control, HT2d, HT4d, and HT6d clusters learned by the VAE and the cluster medians are identified.  
		\textbf{b}, The five most significant peaks for each state are identified. 
        \textbf{c-f}, The Raman spectra are acquired from the leaves of wildtype Choy Sum and Kai Lan plants at ambient temperature of 24 \textsuperscript{o}C (Control) and high temperature of 33 \textsuperscript{o}C at time points as in \textbf{a-b}.
	DIVA applied to wildtype Choy Sum (\textbf{c}, \textbf{d}) and Kai Lan plants (\textbf{e}, \textbf{f}).  
	}
\end{figure}

\begin{figure}
    \centerline{\includegraphics[width=\textwidth]{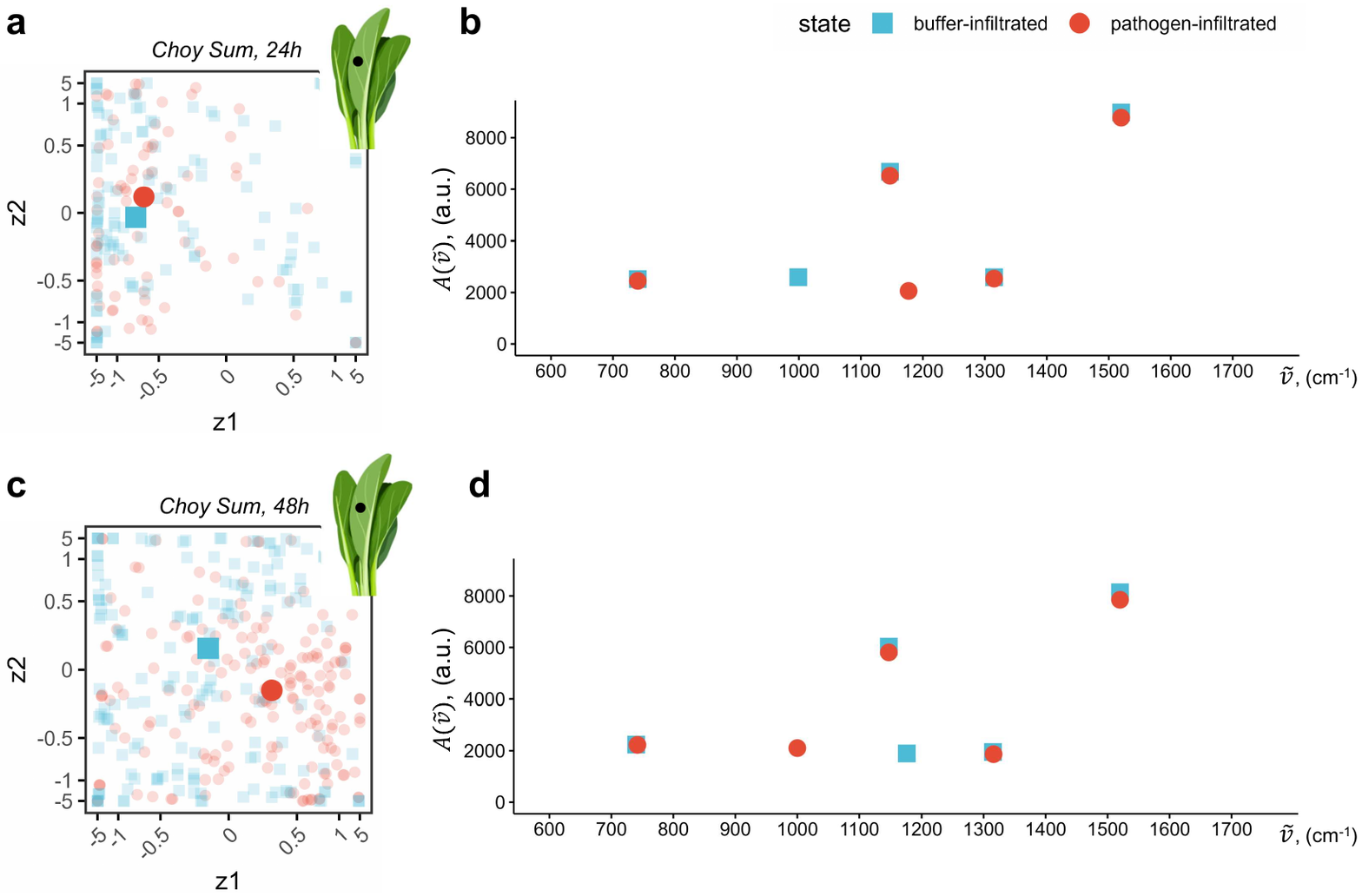}}
    \linenumbers
	\caption{\bodyfigurelabel{Fig5}
		\textbf{Decoding stress responses due to pathogenic bacterial infection in Choy Sum plants.}
        Application of DIVA for pathogenic bacterial stress analysis in leaves of wildtype Choy Sum plants. 
        Refer to ``Bacterial stress study" in Methods for details on pathogenic bacterial stress experiment.
		\textbf{a}, The Raman spectra acquired from the leaves of wildtype buffer-infiltrated and pathogen-infiltrated Choy Sum plants at 24h time point post-infiltration are passed through a VAE for analyzing the plant response to pathogenic bacterial stress at the spectral level. 
		The two-dimensional latent space represents the spectra acquired from the leaves of wildtype buffer-infiltrated and pathogen-infiltrated Choy Sum plants at 24h time point post-infiltration. 
        The 24h clusters learned by the VAE and the cluster medians are identified.  
		\textbf{b}, The five most significant peaks for each group are identified. 
		\textbf{c-d}, DIVA applied to wildtype Choy Sum plants at 48h post-infiltration. 
	}
\end{figure}

\begin{figure}
    \centerline{\includegraphics[width=\textwidth]{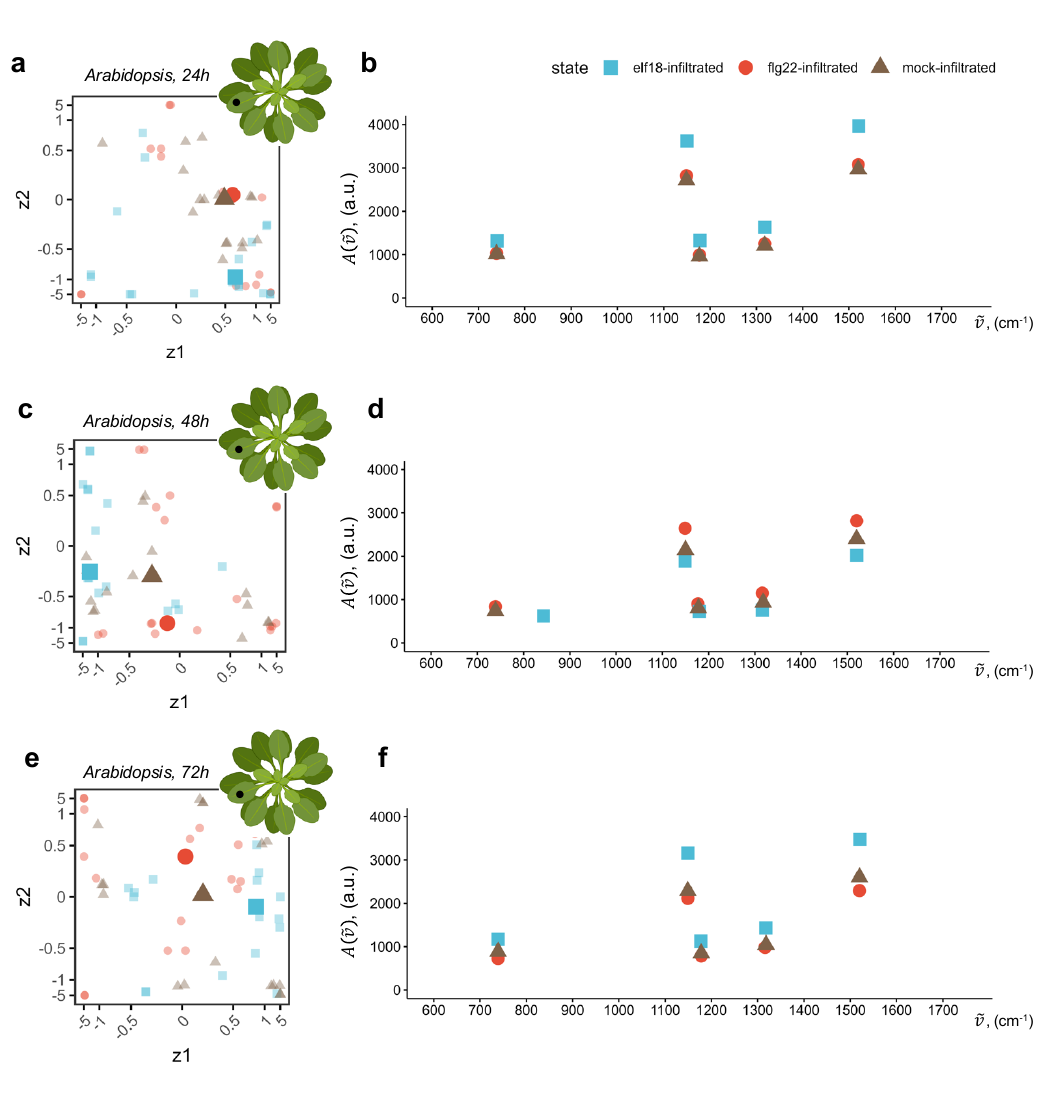}}
    \linenumbers
	\caption{\edfigurelabel{EDFig1}
		\textbf{Decoding stress responses due to elicited bacterial infection in wildtype Arabidopsis plants.}
        Application of DIVA for elicited bacterial stress analysis in leaves of wildtype Arabidopsis plants. 
        \textit{elf18} and \textit{flg22} bacterial elicitors are used in this experiment. 
        Refer to ``Bacterial stress study" in Methods for details on the elicited bacterial stress experiment.
		\textbf{a}, The Raman spectra acquired from the leaves of wildtype mock-infiltrated, \textit{elf18}-infiltrated, and \textit{flg22}-infiltrated Arabidopsis plants at 24h time point post-infiltration are passed through a VAE for analyzing the plant response to elicited bacterial stress at the spectral level. 
		The two-dimensional latent space represents the spectra acquired from the leaves of wildtype mock-infiltrated, \textit{elf18}-infiltrated, and \textit{flg22}-infiltrated Arabidopsis plants at the 24h time point post-infiltration.
        The 24h clusters learned by the VAE and the cluster medians are shown.  
		\textbf{b}, The five most significant peaks for each group are identified. 
		\textbf{c-d}, DIVA applied to wildtype mock-infiltrated, \textit{elf18}-infiltrated, and \textit{flg22}-infiltrated Arabidopsis plants at the 48h time point post-infiltration. 
		\textbf{e-f}, DIVA applied to wildtype mock-infiltrated, \textit{elf18}-infiltrated, and \textit{flg22}-infiltrated Arabidopsis plants at the 72h time point post-infiltration.
	}
\end{figure}

\pagebreak
\begin{methods}
\sloppy	
\noindent\textbf{Raman acquisition system}

The acquisition setup included a benchtop Raman spectrometer\cite{sng2020rapid,chung2021rapid} with an excitation wavelength of 830~nm and laser power of 100~mW.
The 830~nm excitation wavelength was particularly chosen due to its deeper penetration depth and a comparatively lower fluorescence background signal from the plant tissue, resulting in a higher signal-to-noise ratio Raman spectra\cite{sng2020rapid}. 
Raman spectra were acquired in the range 600-1750~cm\textsuperscript{-1}. 
The acquisition time for each of the Raman spectra was 10~s. 
Raman spectra acquired from a polystyrene sample on the day of each experiment were used as a reference to calibrate all the Raman spectra acquired from the plant samples on that particular day. 

\noindent\textbf{Light stress study}
\sloppy

To investigate light stress, 21-day-old soil-grown Arabidopsis, Choy Sum, and Kai Lan plants were measured with Raman spectroscopy. 
The 14-day-old plants were subjected to 7 days of light stress before measurement by Raman spectroscopy. 
All plants were grown under control conditions before exposure to light stress. 
Arabidopsis plants were grown under long-day photoperiod (LD, 16 h light:8 h dark), at 21~\textsuperscript{o}C, and under white light of R:FR=3 and PPFD=100$\,\upmu$mol$\,$m\textsuperscript{-2}s\textsuperscript{-1}. 
Choy Sum and Kai Lan were grown under LD, at 24~\textsuperscript{o}C, and under white light of R:FR=3 and PPFD=150$\,\upmu$mol$\,$m\textsuperscript{-2}s\textsuperscript{-1}.
For Arabidopsis, plants were subjected to various light conditions including white light that served as a control (R:FR=3, PPFD=100$\,\upmu$mol$\,$m\textsuperscript{-2}s\textsuperscript{-1}), high light (R:FR=3, PPFD=300$\,\upmu$mol$\,$m\textsuperscript{-2}s\textsuperscript{-1}), low light (R:FR=3, PPFD=30$\,\upmu$mol$\,$m\textsuperscript{-2}s\textsuperscript{-1}), and shade (R:FR=0.1, PPFD=30$\,\upmu$mol$\,$m\textsuperscript{-2}s\textsuperscript{-1}). 
Likewise, Choy Sum and Kai Lan plants were subjected to light-stress conditions of white light (R:FR=3, PPFD=150$\,\upmu$mol$\,$m\textsuperscript{-2}s\textsuperscript{-1}), high light (R:FR=3, PPFD=600$\,\upmu$mol$\,$m\textsuperscript{-2}s\textsuperscript{-1}), low light (R:FR=3, PPFD=30$\,\upmu$mol$\,$m\textsuperscript{-2}s\textsuperscript{-1}), and shade (R:FR=0.1, PPFD=30$\,\upmu$mol$\,$m\textsuperscript{-2}s\textsuperscript{-1}). 
White light was supplied by SE-004 4000K white light LED tubes (Systion Electronics, Portugal). 
Low R:FR was achieved by supplementing low-intensity white light with far-red light from SE-004 FAR RED-730nm LED tubes (Systion Electronics, Portugal).
Raman spectra were acquired on the leaf blades for all plant samples, specifically on the third true leaf for Arabidopsis plants and the first true leaf for Choy Sum and Kai Lan plants, respectively. 
Two acquisitions were made per leaf blade (one on each side of the main vein).

\noindent\textbf{Shade avoidance stress study}

For the case study on shade-avoidance stress, 17-day-old Arabidopsis plants grown in soil were measured with Raman spectroscopy. 
Mutant plants \textit{phyA-211} and \textit{phyB-9\textsuperscript{BC}} were included along with the wildtype Arabidopsis plants for analysis.
10-day-old plants were subjected to various growth conditions for 7 days, including white light that served as a control (WL, red:far-red ratio (R:FR) of 3 and photosynthetic photon flux density (PPFD) of 100$\,\upmu$mol$\,$m\textsuperscript{-2}s\textsuperscript{-1}), moderate shade (R:FR of 0.7 and PPFD of 60$\,\upmu$mol$\,$m\textsuperscript{-2}s\textsuperscript{-1}), and deep shade (R:FR of 0.2 and PPFD of 30$\,\upmu$mol$\,$m\textsuperscript{-2}s\textsuperscript{-1}). 
Raman spectra were acquired on the third true leaf, on the leaf blade, and on the petiole. 
Two acquisitions were made per leaf blade (one on each side of the main vein) and one acquisition was made for the petiole (at the middle of the petiole). 
More details regarding the plant samples appear in our previous study\cite{sng2020rapid}. 

\noindent\textbf{High-temperature stress study}

For the high-temperature stress study, 21-day-old soil-grown Arabidopsis, Choy Sum, and Kai Lan plants were measured with Raman spectroscopy. 
Arabidopsis, Choy Sum, and Kai Lan plants were grown in the control condition (21~\textsuperscript{o}C for Arabidopsis and 24~\textsuperscript{o}C for Choy Sum and Kai Lan) until subjected to their respective high temperature treatments. 
Arabidopsis plants were subjected to high ambient temperature at 29~\textsuperscript{o}C, while Choy Sum and Kai Lan were exposed to 33~\textsuperscript{o}C. 
The plants were subjected to various high-temperature durations (0, 2, 4, 6 days) before data acquisition. 
All plants were of the same age during data acquisition. 
Throughout the experiment, all plants were grown under LD photoperiod and white light condition (Arabidopsis: R:FR=3, PPFD=100$\,\upmu$mol$\,$m\textsuperscript{-2}s\textsuperscript{-1}; Choy Sum and Kai Lan: R:FR=3, PPFD=150$\,\upmu$mol$\,$m\textsuperscript{-2}s\textsuperscript{-1}).
Raman spectra were acquired on the leaf blade and the corresponding petiole of all the plant samples, specifically on the third true leaf for Arabidopsis plants, and the first true leaf for Choy Sum and Kai Lan plants. 
The data acquisitions for the stressed states were made on days 2, 4, and 6 post-induction of high-temperature stress. 
Five Raman spectra were obtained per acquisition. 
Two acquisitions were made per leaf blade (one on each side of the main vein) and one acquisition was made per petiole (middle of the petiole). 

\noindent\textbf{Bacterial stress study}

For the case study on bacterial stress, Choy Sum plants for pathogen infection and Arabidopsis plants for bacterial elicitation were considered. 

Commercial Choy Sum ({\it Brassica rapa} var. {\it parachinensis}) seeds were germinated and grown under controlled environmental conditions (25~\textsuperscript{o}C, 75\% humidity, 16h light/8h dark photoperiod, 230$\,\upmu$mol$\,$m\textsuperscript{-2}s\textsuperscript{-1} light intensity). 
\textit{Xanthomonas campestris} pv.~\textit{campestris} (Xcc) isolate (ATCC33913) was obtained from the American Type Culture Collection and cultured in YGC medium (yeast extract, glucose, calcium carbonate) at 28~\textsuperscript{o}C in a shaker incubator at 125 rpm. 
Bacterial suspensions were prepared by resuspending cultures in fresh YGC to a final concentration of 1×10$^8$ CFU mL\textsuperscript{-1}.
Two-week-old plants were used for infiltration. 
The abaxial side of the fourth true leaf was inoculated with bacterial suspension using a needleless syringe on either side of the central vein near the apex margins. 
Control plants were mock-inoculated with sterile YGC medium. 
All plants were maintained under identical growth conditions following treatment.
Leaf discs ($\sim$5 mm diameter) were collected from the proximal region of the infiltrated zone at 24~h and 48~h post-infiltration. 
From each disc, three positions were measured, acquiring five Raman spectra per position (10~s exposure per spectrum). 
Thirty spectra per biological replicate were averaged to generate a single representative Raman profile. 
Three biological replicates were collected for each treatment and time point.

\sloppy
Arabidopsis plants were grown at 21 \textsuperscript{o}C for 6 weeks and challenged with \textit{elf18} and \textit{flg22} bacterial elicitors.
More details regarding the plant growth and elicitor/pathogen infiltration appear in our previous study\cite{chung2021rapid}. 
Raman spectra were acquired from the leaf blade discs harvested from the infiltrated leaf in all the plant samples, specifically on the seventh or the eighth leaf for Arabidopsis plants. 
The data acquisitions for the stressed states were made at 24~h, 48~h, and 72~h post-infiltration for the case of infection with bacterial elicitors.

\noindent\textbf{Data preprocessing}

Raw Raman spectra (I($\tilde{v}$)) acquired via benchtop systems and, in some experiments, via portable systems, were directly fed into the workflow as detailed in Fig. \ref{Fig1}. 
No feature selection or meticulous data preprocessing steps inclusive of smoothening and baseline removal were explicitly performed for any of the data, as it would bias the training of the autoencoder. 
Spectral trimming below 600~cm\textsuperscript{-1} and above 1750~cm\textsuperscript{-1} to exclude amplified emission effects caused by the incident laser and removal of cosmic-ray noise were performed using \textit{BRAPH2DiSTAP} software, a custom fork of the BRAPH 2.0 software\cite{Chang2025.04.11.648455,mijalkov2017braph}. 
The \textit{BRAPH2DiSTAP} software was developed and compiled in MATLAB\textsuperscript{\textregistered} v2023b. 
The spectra were normalized by reducing the original Raman intensities because the values were too large for the neural network to train effectively.
Without normalization, the high intensities at the start of the wavenumber range would have caused an excessively large loss at the beginning of training, making it harder for the model to converge.
A first derivative transformation was then applied to enable the encoder to capture relative trends in Raman intensity changes. 
\begin{equation}
D(\tilde{v}) = \frac{dI(\tilde{v})}{d\tilde{v}}
\end{equation}
\noindent where $D(\tilde{v})$ is the first derivative representation of an original Raman spectrum $I(\tilde{v})$, which serves as a representative input to the encoder. 
The total dataset is then randomly divided, with $\sim$ 90\% allocated for training and $\sim$10\% for testing, ensuring a robust evaluation of the VAE model performance.

\noindent\textbf{Variational Autoencoder for Unsupervised Clustering}

In this study, we employed a variational autoencoder (VAE)\cite{kingma2019introduction,doersch2016tutorial} to perform unsupervised clustering based on local microenvironments, characterized by distinct Raman spectral descriptors. 
These inherently latent descriptors were derived from A($\tilde{v}$), which was evaluated at each peak position of the characteristic Raman spectra derivative D($\tilde{v}$). 
The VAE architecture enabled the generation of a continuous and structured latent space of Raman spectral features, where each point corresponded to a unique spectral signature\cite{dlcc}.

The VAE consisted of two main components: the encoder and the decoder. 
The encoder network began with an input layer whose size matched the length of the input derivative spectrum D($\tilde{v}$). 
This was followed by a fully connected layer with 64 neurons and a Rectified Linear Unit (ReLU) activation layer. 
A subsequent fully connected layer with 4 neurons fed into the sampling layer, concluding the encoder. 
The decoder network mirrored this structure with two fully connected layers and included a Leaky ReLU activation layer with an alpha value of 1, allowing linear activation for both positive and negative inputs.
The sampling layer processed two vectors: the mean vector $\mu$ and the log-variance vector $\log(\sigma^{2})$. 
Elements were sampled from a normal distribution $N(\mu_{i}, \sigma_{i}^{2})$ using these vectors, with $\log(\sigma^{2})$ included to enhance numerical stability during training. 
The VAE was trained to enforce a Gaussian distribution on the latent space, encouraging smoothness and continuity in the learned representation.
The training process used a batch size of 40 over 3000 epochs with a learning rate of 10\textsuperscript{-3}. 
Optimization was driven by the Evidence Lower Bound (ELBO) loss function, which balances reconstruction accuracy and latent space regularization. 
The ELBO loss comprises two main components: the reconstruction loss and the Kullback–Leibler (KL) divergence loss.
The reconstruction loss was computed using the mean absolute error (MAE) between the decoder’s output $\widehat{x}{i}$ and the original input $x{i}$, as shown:
\begin{equation}
\text{reconstruction loss} = \frac{1}{N}\sum_{i=1}^{N}|x_{i} - \widehat{x}_{i}|
\end{equation}
\noindent where $N$ represents the number of input variables.

The KL divergence loss measured the deviation of the learned latent variable distribution from the standard normal distribution, ensuring that the latent space remained regularized and generative:
\begin{equation}
\text{KL loss} = -\frac{1}{2}\sum_{i=1}^{M}\left( 1+log(\sigma_{i}^{2})-\mu_{i}^{2}-\sigma_{i}^{2}) \right)
\end{equation}
\noindent where $M$ is the number of latent variables.

Together, these loss components formed the ELBO loss function, which was minimized to train the VAE. 
Including the KL divergence term helped ensure that the clusters formed through the reconstruction loss were tightly concentrated around the center of the latent space. 
This resulted in a continuous and well-defined space for sampling and interpretation.

\noindent\textbf{Training set sizes and Model execution times}

For each case study, the DIVA model was trained on Raman spectral data collected under different stress treatments and plant genotypes. 
To ensure robust evaluation, 90\% of the data was used for training, and results presented in the main figures are based on these training sets. 
The remaining 10\% was held out for testing and validation.

\noindent\textbf{1. Light stress study}
This dataset included three species---Arabidopsis, Choy Sum, and Kai Lan---exposed to four light conditions: high light, low light, white light, and deep shade. 
Spectra contained 871 Raman features.
\textit{Arabidopsis}: 160 spectra from 32 locations (5 replicates per location), divided equally into 8 locations per light condition (4 leaves per condition, 2 locations per leaf).
\textit{Choy Sum}: 150 spectra from 30 locations, with 8 locations each for high, low, and white light, and 6 for deep shade (2 locations per leaf).
\textit{Kai Lan}: 160 spectra from 32 locations, divided equally across the 4 light conditions (8 locations per condition, 2 locations per leaf).
Training time: $\sim$3–4 minutes.

\noindent\textbf{2. Shade avoidance stress study}
This dataset focused on Arabidopsis genotypes responding to shade. 
Each spectrum had 950 features.
\textit{phyA-211 group}: 100 spectra from 20 locations (5 replicates per location), with 6, 8, and 6 locations under deep shade, mild shade, and white light respectively (2 locations per leaf).
\textit{phyB-9\textsuperscript{BC} group}: 60 spectra from 12 locations, equally divided across 3 light conditions.
\textit{Wildtype group}: 60 spectra from 12 locations, divided similarly across conditions (2 locations per leaf).
Training time: $\sim$2–3 minutes per model.

\noindent\textbf{3. High-temperature stress study}
This study examined thermal stress in three species. 
Each spectrum had 871 Raman features.
\textit{Arabidopsis}: 120 spectra from 24 locations, with 6 locations per temperature condition (4 groups total).
\textit{Choy Sum}: 80 spectra from 16 locations, with 4 per condition.
\textit{Kai Lan}: 80 spectra from 16 locations, with 4 per condition.
Training time: $\sim$2–3 minutes per model.

\noindent\textbf{4. Bacterial stress study (Pathogen-induced infection)}
This dataset analyzed Choy Sum leaves infiltrated with a bacterial suspension or buffer. 
Each spectrum contained 839 Raman features.
\textit{24 hours post-infiltration}: 225 spectra from 45 locations (5 replicates per location), with 25 for buffer and 20 for infected group.
\textit{48 hours post-infiltration}: 400 spectra from 80 locations, equally divided into 40 locations per group.
Training time: $\sim$3–4 minutes per time point.

\noindent\textbf{5. Bacterial stress study (Elicitor-induced infection)}
This dataset involved Arabidopsis wild-type plants infiltrated with bacterial elicitors. 
Each spectrum contained 823 Raman features.
\textit{Wildtype control group}: 180 spectra from 36 locations (5 replicates per location), with 12 locations per time point.
\textit{elf18-infiltrated group}: 180 spectra from 36 locations, with 12 per time point.
\textit{flg22-infiltrated group}: 180 spectra from 36 locations, with 12 per time point.
Training time: $\sim$3 minutes per group.

\noindent\textbf{Model Execution Time}
Once trained, the DIVA model processed each study---including significant peak detection, latent space projection, and visualization---in approximately 3–5 seconds.

\noindent\textbf{Significant Peaks Detection Methodology}

The detection of significant peaks in the stressed state of a plant was achieved using a zero-crossing analysis implemented in MATLAB. 
The process identified mainly two types of zero-crossing events within the input signal array D($\tilde{v}$), sampled at corresponding indices: positive-to-negative zero-crossings and general zero-crossings. 
For each type, crossing points were computed using linear interpolation at a zero threshold. 
Specifically, the function for positive-to-negative zero-crossings returned indices (\textit{ind\_p2n}), interpolated crossing times (\textit{t0\_p2n}), and signal values at the crossing points (\textit{s0\_p2n}). 
Similar computations were performed for all zero-crossings resulting in \textit{ind}, \textit{t0}, \textit{s0} respectively.

To quantify the significance of peaks, the area under the graph surrounding each positive-to-negative zero-crossing was calculated. 
For each such crossing index, the area was computed as the sum of the absolute signal values between the immediately preceding and succeeding zero-crossing points.
To handle boundary conditions, incomplete peak regions at the start and end of the spectrum that lacked valid neighboring zero-crossings were discarded, ensuring only complete and interpretable peak regions were considered.
The resulting area values were stored in an array, representing an estimate of the biomolecular concentration at the site of Raman signal acquisition.

The identified peaks were compiled into a matrix, \textit{sig\_pks}, which included the zero-crossing times (\textit{t0\_p2n}), their corresponding rounded indices, and the computed area values. 
For ranking purposes, the peaks were sorted in descending order based on their area-under-the-graph values. 
This ranking yielded a matrix containing the sorted zero-crossing times, rounded indices, and area-under-the-graph values. 
This final step facilitated the prioritization of peaks, emphasizing the most significant features of the stressed state of the plant.

\noindent\textbf{Hardware for accelerated evaluation}

An Nvidia RTX 4080 was used for the GPU workload while training the variational auto-encoder models. 
The CPU used is an Intel\textsuperscript{\textregistered} Core\textsuperscript{TM} 14th Gen i9 14900KF. 
64GB DDR5 RAM was employed for all model calculations pertinent to training and evaluation. 
The training was performed on this set of hardware running Windows 11 Pro (x64) with no other foreground programs running when training was conducted. 

\end{methods}


\pagebreak
\noindent{\bfseries References}\setlength{\parskip}{12pt}%
\bibliography{p1}

\pagebreak
\begin{addendum}
\sloppy
\item [Acknowledgments] The study was supported by (i) Disruptive \& Sustainable Technologies for Agricultural Precision (DiSTAP), Singapore-MIT Alliance for Research and Technology (SMART),  (ii) Temasek Life Sciences Laboratory (TLL), National University of Singapore (NUS) and (iii) the National Research Foundation (NRF). The computations and data handling were enabled by the resources in project NAISS 2023/22-349 provided by the National Academic Infrastructure for Supercomputing in Sweden (NAISS), partially funded by the Swedish Research Council through grant agreement no.~ 2022-06725. Data curation for deep learning was carried out using \textit{BRAPH2DiSTAP} software, a customized fork of the BRAPH 2.0 software (https://github.com/braph-software/BRAPH-2). The authors acknowledge the original contributions to BRAPH 2.0 that have helped shape further development in the form of data analysis pipelines used in this work. The authors thank the researchers from the Genome \& Ecological Biology, Molecular Pathogenesis, and Developmental Biology groups at TLL for providing the Raman spectroscopy data for deep learning tasks reported in this work and for their valuable support throughout the study. The graphical schematics related to plant types were created in part using BioRender.com. 

\item[Author Contributions] G.V. and G.P.S. conceived the research project and collaborated with I.-C.J., R.S., and C.N.-H. for this work. G.V., A.C.P., Y.-W.C., and G.P.S. conceived the technique reported in this manuscript. G.V. and A.C.P. conceived the design and content of this manuscript. A.C.P., Y.-W.C., G.V., and J.B.P. curated the data. A.C.P. and Y.-W.C. performed all analyses reported in this manuscript. B.J.R.S., I.-C.J., R.S., and C.N.-H. designed and conducted the experiments. A.C.P. and G.V. wrote the manuscript with input from all authors. All authors have approved the final version of the manuscript.

\item[Competing Interests] The authors declare they have no competing financial interests.

\item[Correspondence] Correspondence and information requests should be addressed to {\it giovanni.volpe@physics.gu.se},  {\it jangi@tll.org.sg}, or {\it gajendra@smart.mit.edu}.
\end{addendum}

\end{document}


\sloppy
\noindent \Large\textbf{Deep-Learning Investigation of Vibrational Raman Spectra for Plant-Stress Analysis}
\normalsize
\vspace{10mm}\\
\noindent \large\textbf{Supplementary Material}
\vspace{10mm}\\
\normalsize
\noindent Anoop C. Patil\textsuperscript{1*}, Benny Jian Rong Sng\textsuperscript{1,2*}, Yu-Wei Chang\textsuperscript{3*}, Joana B. Pereira\textsuperscript{4}, Chua Nam-Hai\textsuperscript{1,2}, Rajani Sarojam\textsuperscript{1,2}, Gajendra Pratap Singh\textsuperscript{1+}, In‑Cheol Jang\textsuperscript{1,2,5+}, and Giovanni Volpe\textsuperscript{1,3,6+}
\vspace{8mm}

\noindent \textsuperscript{1} \textit{Disruptive \& Sustainable Technologies for Agricultural Precision, 1 CREATE way, Singapore-MIT Alliance for Research and Technology, Singapore 138602, Singapore.}

\noindent \textsuperscript{2} \textit{Temasek Life Sciences Laboratory, 1 Research Link, National University of Singapore, Singapore 117604, Singapore.}

\noindent \textsuperscript{3} \textit{Department of Physics, University of Gothenburg, Gothenburg 41296, Sweden.}

\noindent \textsuperscript{4} \textit{Department of Clinical Neuroscience, Karolinska Institute, Stockholm 17165, Sweden.}

\noindent \textsuperscript{5} \textit{Department of Biological Sciences, National University of Singapore, Singapore 117543, Singapore.}

\noindent \textsuperscript{6} \textit{Science for Life Laboratory, Department of Physics, University of Gothenburg, Gothenburg 41296, Sweden.}

\noindent \vspace{5mm}\\
$^{*}$\textit{Authors contributed equally to this work}\\
$^{+}$\textit{To whom correspondence should be addressed:\\giovanni.volpe@physics.gu.se; jangi@tll.org.sg; gajendra@smart.mit.edu}

\normalsize
\pagebreak

\small
\begin{spacing}{1.25}
\tableofcontents
\end{spacing}
\newpage

\begin{spacing}{1.25}
\vspace{-5mm}

\begingroup
\sloppy
\listoffigures
\endgroup

\newpage

\begingroup
\sloppy
\listoftables
\endgroup

\end{spacing}
\pagebreak
\normalsize
	
\section{Supplementary Figures}

\begin{figure}[htbp]
	\centering
    \includegraphics[width=\textwidth]{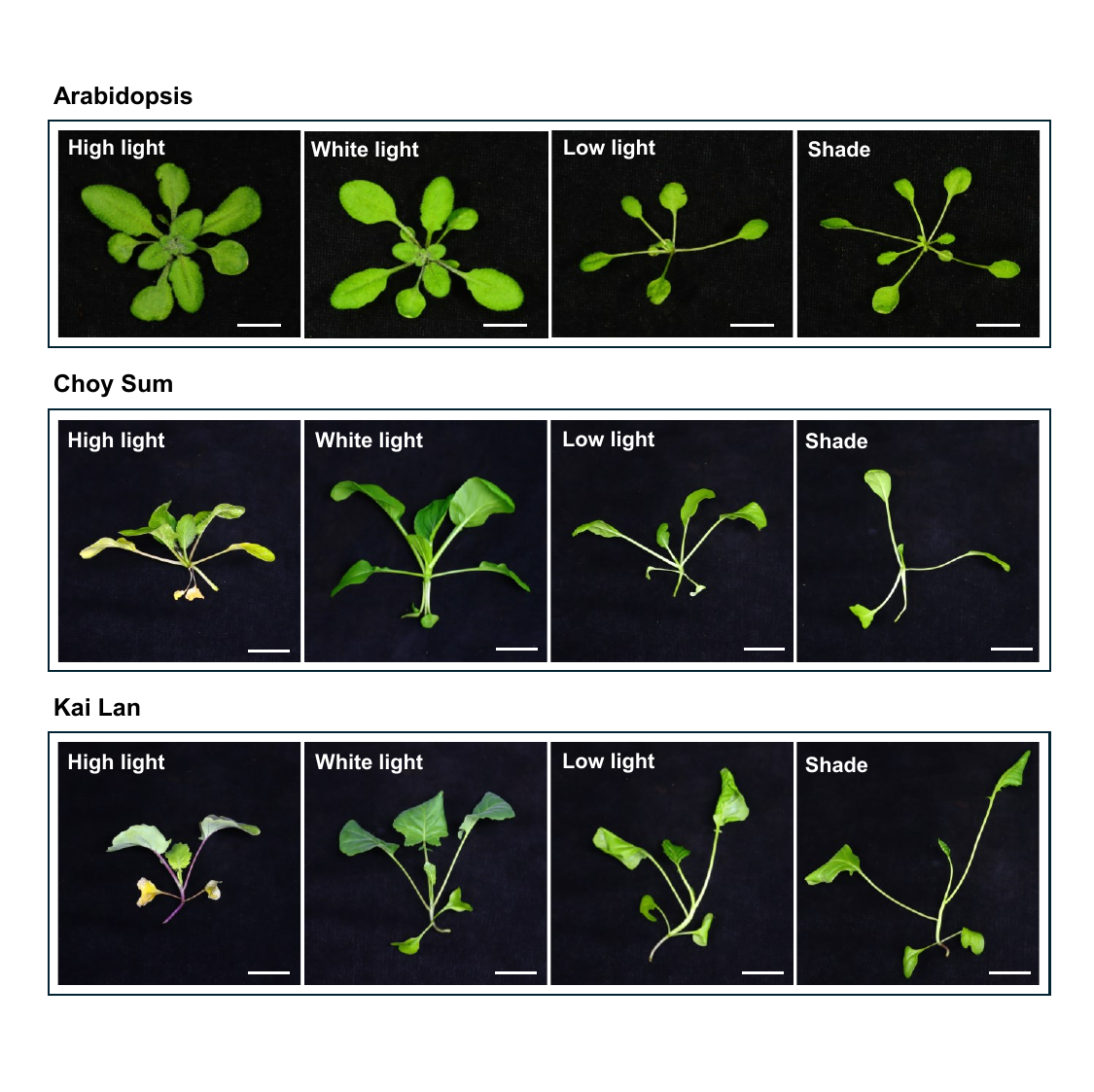} 
	\caption[Phenotype of wildtype Arabidopsis, Choy Sum, and Kai Lan grown under various light stresses.]{\textbf{Phenotype of wildtype Arabidopsis, Choy Sum, and Kai Lan grown under various light stresses.} All the wildtype plants were grown under white light for 2 w, followed by light stress treatment (white light, high light, low light, or shade conditions) for an additional 1 w. For Arabidopsis plant images ---Scale, 1 cm. For Choy Sum and Kai Lan images --- Scale, 3 cm.} 
	\label{SFig1}      
\end{figure}
\clearpage

\begin{figure}[htbp]
	\centering
	\includegraphics[width=\textwidth]{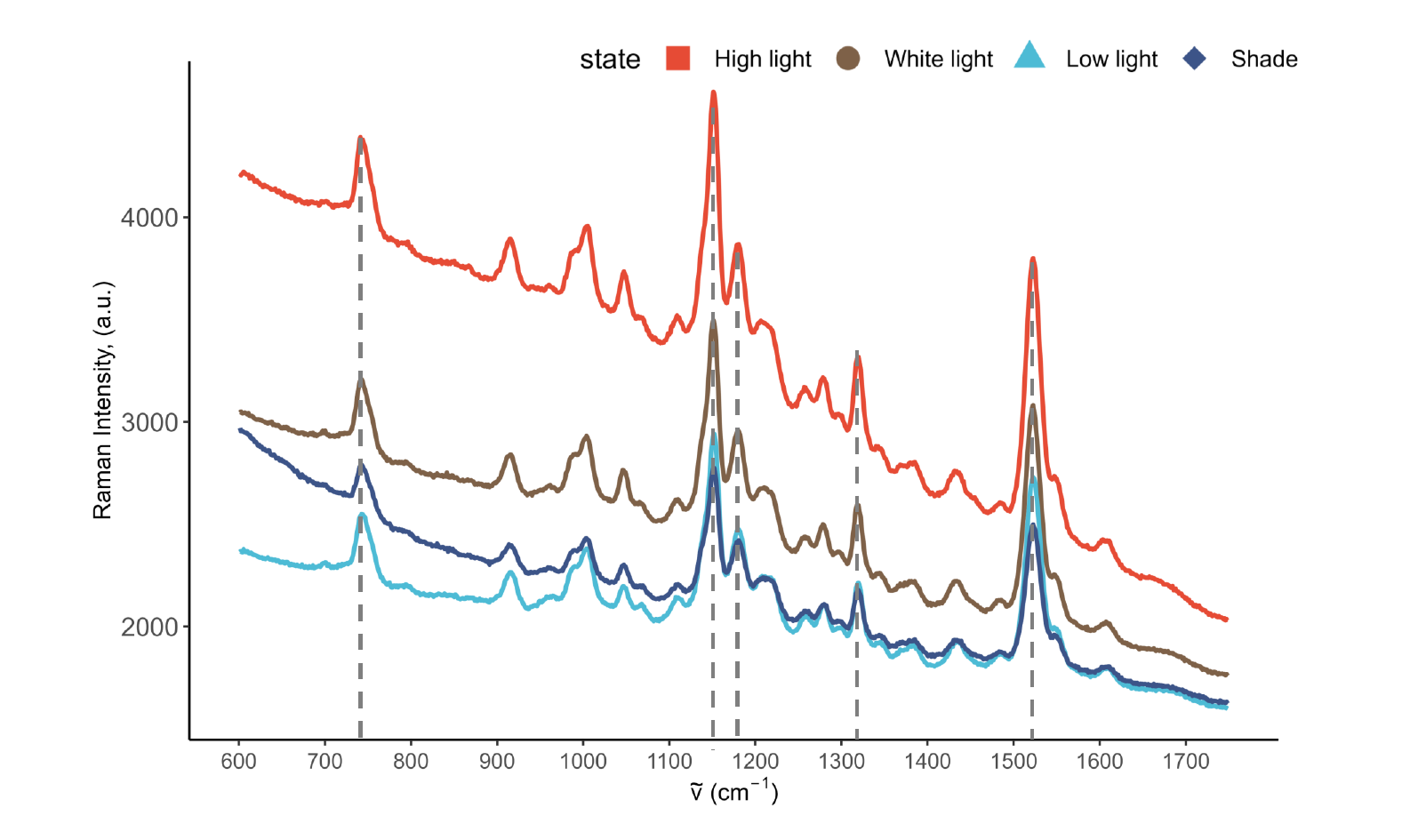} 
	\caption[Visualization of the original Raman spectra representing the light stress conditions in wildtype Arabidopsis plants.]{\textbf{Visualization of the original Raman spectra representing the light stress conditions in wildtype Arabidopsis plants.} The original spectra were obtained by detransforming the derivative D($\tilde{v}$) spectra (shown in Fig.~2b) reconstructed from the cluster medians of Arabidopsis data in Fig.~2a. The five most significant peaks for each light condition are highlighted with dotted lines.} 
	\label{SFig2}  
\end{figure}
\clearpage

\begin{figure}[htbp]
	\centering
	\includegraphics[width=\textwidth]{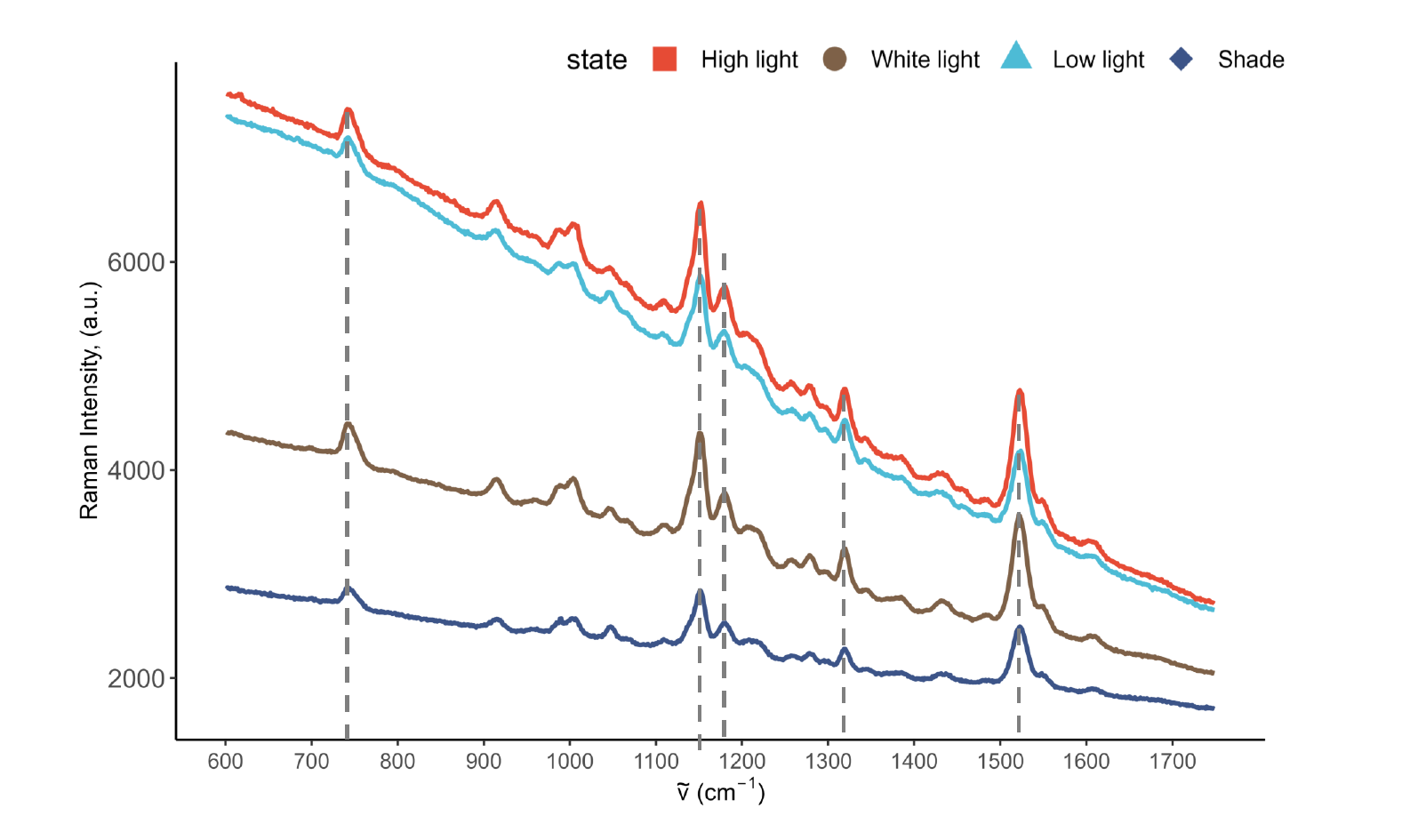} 
	\caption[Visualization of the original Raman spectra representing the light stress conditions in wildtype Choy Sum plants.]{\textbf{Visualization of the original Raman spectra representing the light stress conditions in wildtype Choy Sum plants.} The original spectra were obtained by detransforming the derivative D($\tilde{v}$) spectra reconstructed from the cluster medians of Choy Sum data in Fig.~2c. The five most significant peaks for each light condition are highlighted with dotted lines.} 
	\label{SFig3}      
\end{figure}
\clearpage

\begin{figure}[htbp]
	\centering
	\includegraphics[width=\textwidth]{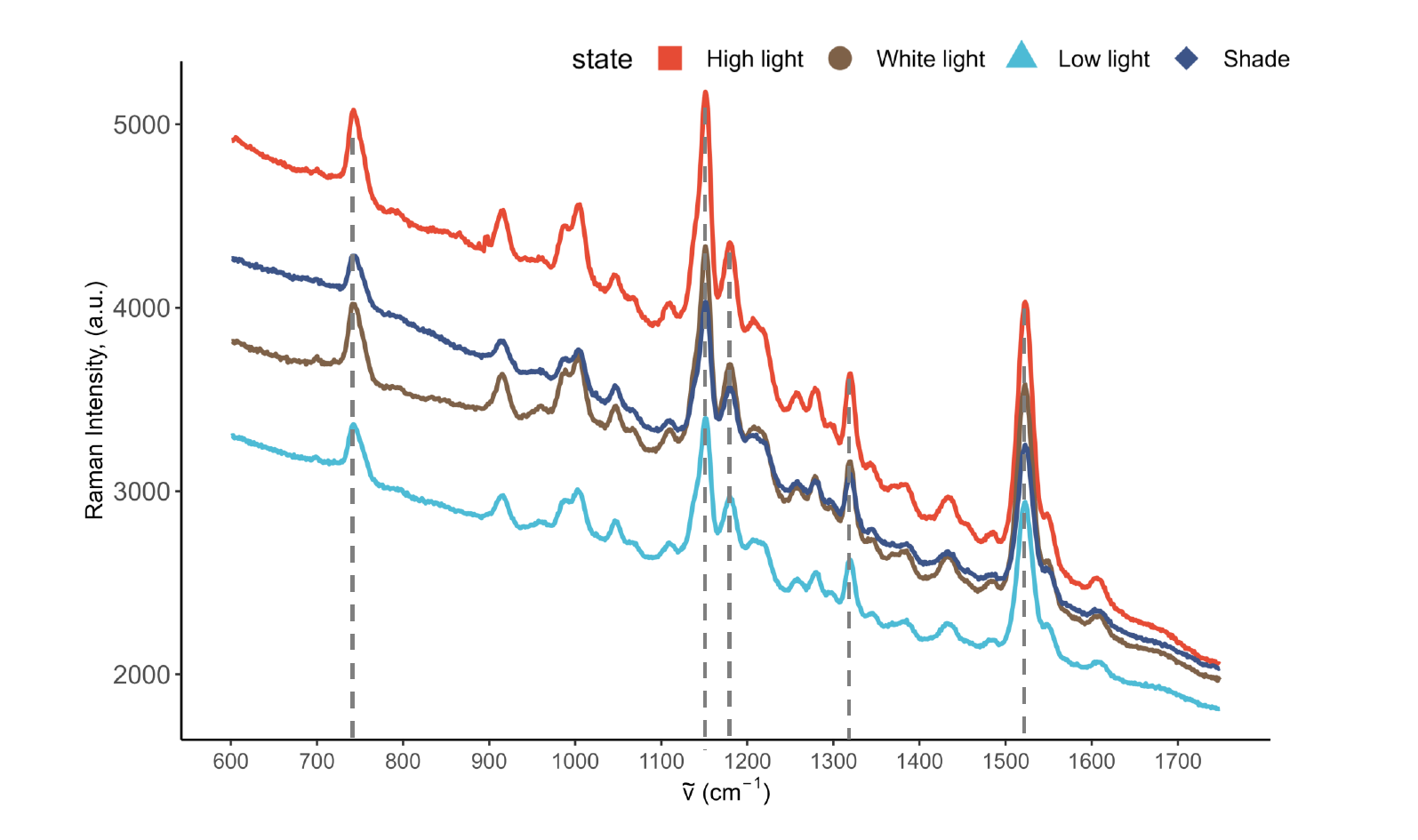} 
	\caption[Visualization of the original Raman spectra representing the light stress conditions in wildtype Kai Lan plants.]{\textbf{Visualization of the original Raman spectra representing the light stress conditions in wildtype Kai Lan plants.} The original spectra were obtained by detransforming the derivative D($\tilde{v}$) spectra reconstructed from the cluster medians of Kai Lan data in Fig.~2e. The five most significant peaks for each light condition are highlighted with dotted lines.} 
	\label{SFig4}      
\end{figure}
\clearpage

\begin{figure}[htbp]
	\centering
    \includegraphics[width=\textwidth]{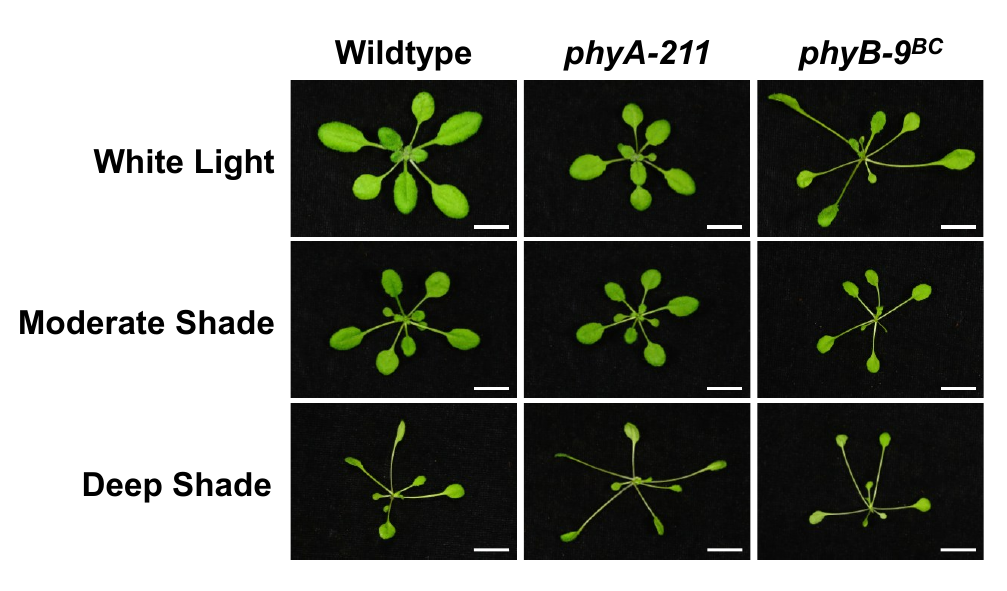} 
	\caption[Phenotype of Arabidopsis plants undergoing shade avoidance response.]{\textbf{Phenotype of Arabidopsis plants undergoing shade avoidance response.} Wildtype, \textit{phyA-211} and \textit{phyB-9\textsuperscript{BC}} Arabidopsis plants were grown under white light for 2 w, followed by shade treatment (white light, mild-shade or deep-shade) for 1 w. WT, wildtype. WL, white light. MS, mild shade. DS, deep shade. Scale, 1 cm.} 
	\label{SFig5}      
\end{figure}
\clearpage

\begin{figure}[htbp]
	\centering
    \includegraphics[width=\textwidth]{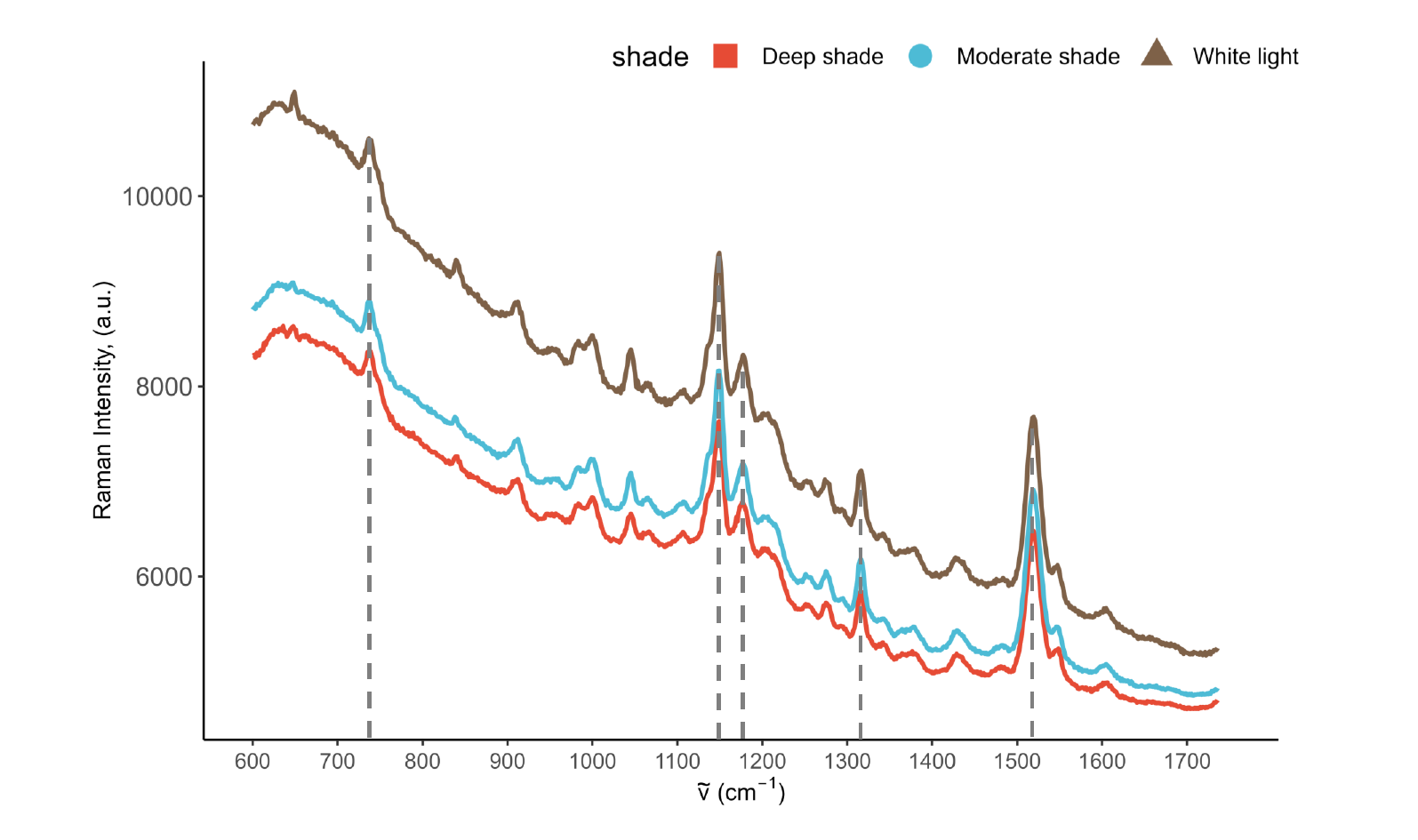} 
	\caption[Visualization of the original Raman spectra representing the shade avoidance stress conditions in wildtype Arabidopsis leaves.]{\textbf{Visualization of the original Raman spectra representing the shade avoidance stress conditions in wildtype Arabidopsis leaves.} The original spectra were obtained by detransforming the derivative D($\tilde{v}$) spectra reconstructed from the cluster medians of wildtype Arabidopsis leaf data in Fig.~3a. The five most significant peaks for each shade condition are highlighted with dotted lines.}
	\label{SFig6}      
\end{figure}
\clearpage

\begin{figure}[htbp]
	\centering
    \includegraphics[width=\textwidth]{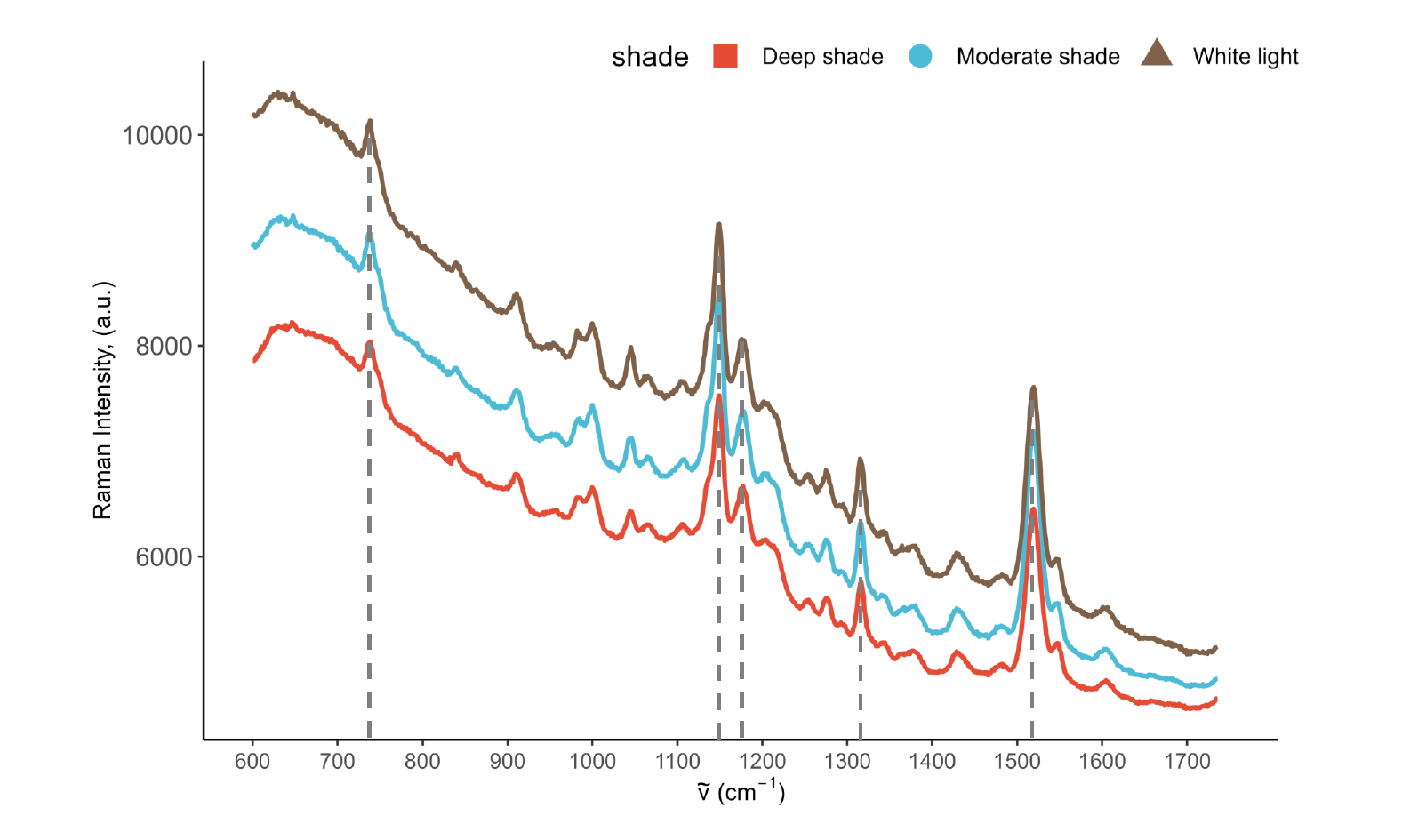} 
	\caption[Visualization of the original Raman spectra representing the shade avoidance stress conditions in \textit{phyA-211} Arabidopsis leaves.]{\textbf{Visualization of the original Raman spectra representing the shade avoidance stress conditions in \textit{phyA-211} Arabidopsis leaves.} The original spectra were obtained by detransforming the derivative D($\tilde{v}$) spectra reconstructed from the cluster medians of \textit{phyA-211} Arabidopsis leaf data in Fig.~3c. The five most significant peaks for each shade condition are highlighted with dotted lines.} 
	\label{SFig7}      
\end{figure}
\clearpage

\begin{figure}[htbp]
	\centering
    \includegraphics[width=\textwidth]{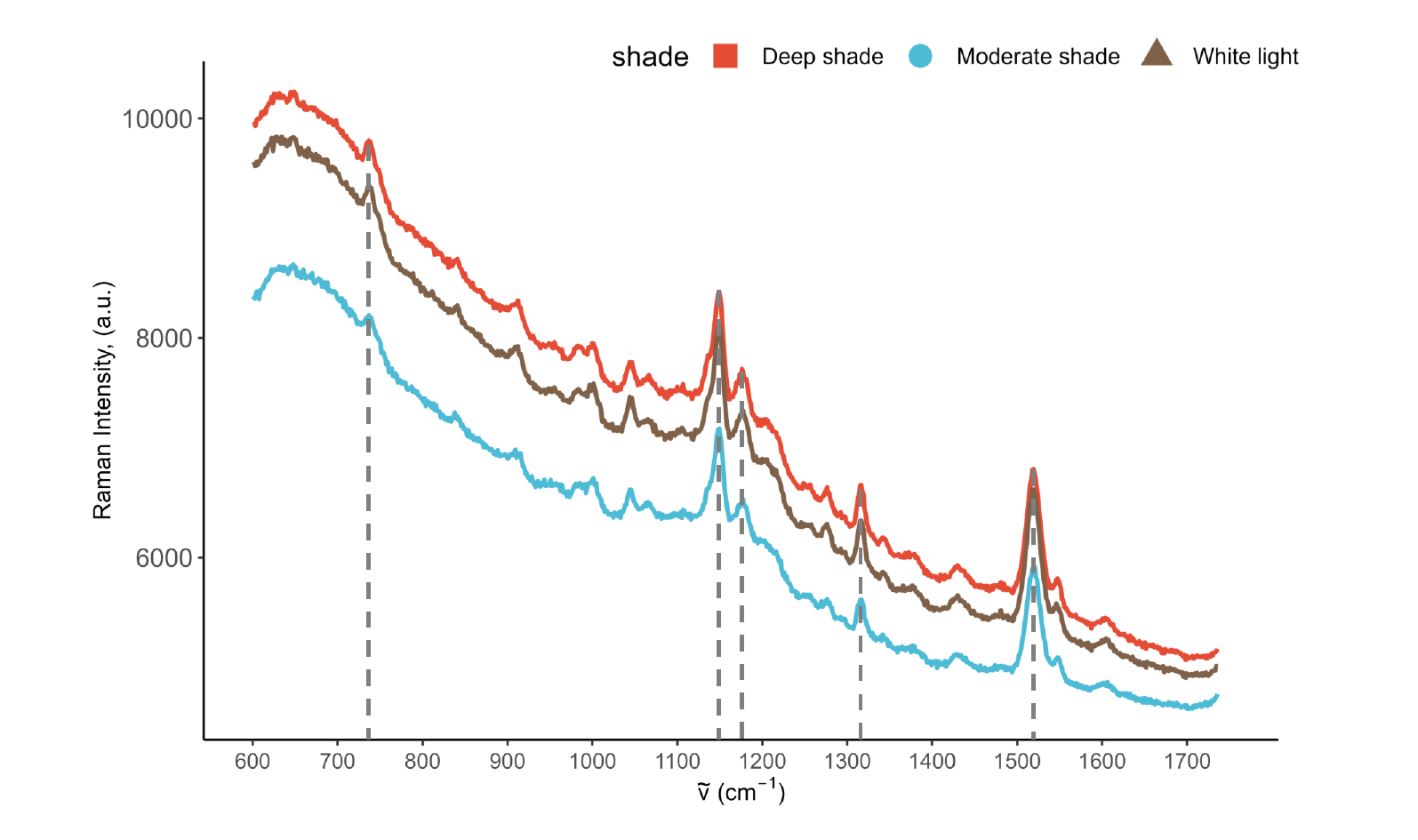} 
	\caption[Visualization of the original Raman spectra representing the shade avoidance stress conditions in \textit{phyB-9\textsuperscript{BC}} Arabidopsis leaves.]{\textbf{Visualization of the original Raman spectra representing the shade avoidance stress conditions in \textit{phyB-9\textsuperscript{BC}} Arabidopsis leaves.} The original spectra were obtained by detransforming the derivative D($\tilde{v}$) spectra reconstructed from the cluster medians of \textit{phyB-9\textsuperscript{BC}} Arabidopsis leaf data in Fig.~3e. The five most significant peaks for each shade condition are highlighted with dotted lines.} 
	\label{SFig8}      
\end{figure}
\clearpage

\begin{figure}[htbp]
	\centering
    \includegraphics[width=\textwidth]{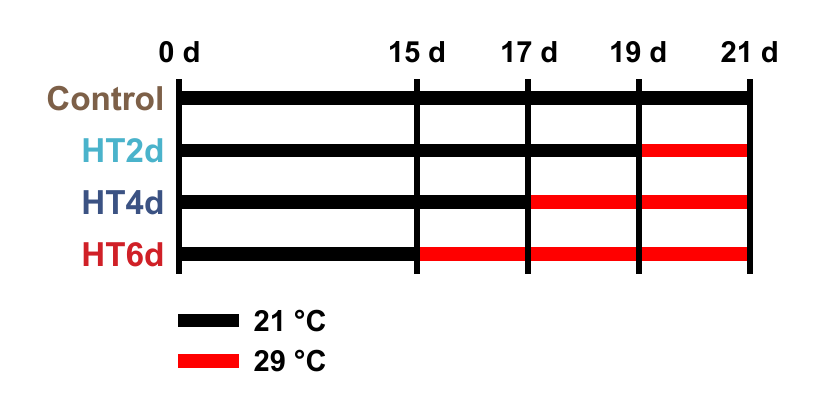} 
	\caption[Schematic diagram illustrating different durations of high-temperature treatment used for wildtype Arabidopsis.]{\textbf{Schematic diagram illustrating different durations of high-temperature treatment used for wildtype Arabidopsis.} Wildtype Arabidopsis were grown at 21 \textsuperscript{o}C before exposure to high temperature (29 \textsuperscript{o}C) over various durations. Control: 21 \textsuperscript{o}C, HT2d: Day-2 time point at high temperature (29 \textsuperscript{o}C), HT4d: Day-4 time point at high temperature (29 \textsuperscript{o}C), and HT6d: Day-6 time point at high temperature (29 \textsuperscript{o}C). All plants were 3 w old at the end of the treatment.} 
	\label{SFig9}      
\end{figure}
\clearpage

\begin{figure}[htbp]
	\centering
    \includegraphics[width=\textwidth]{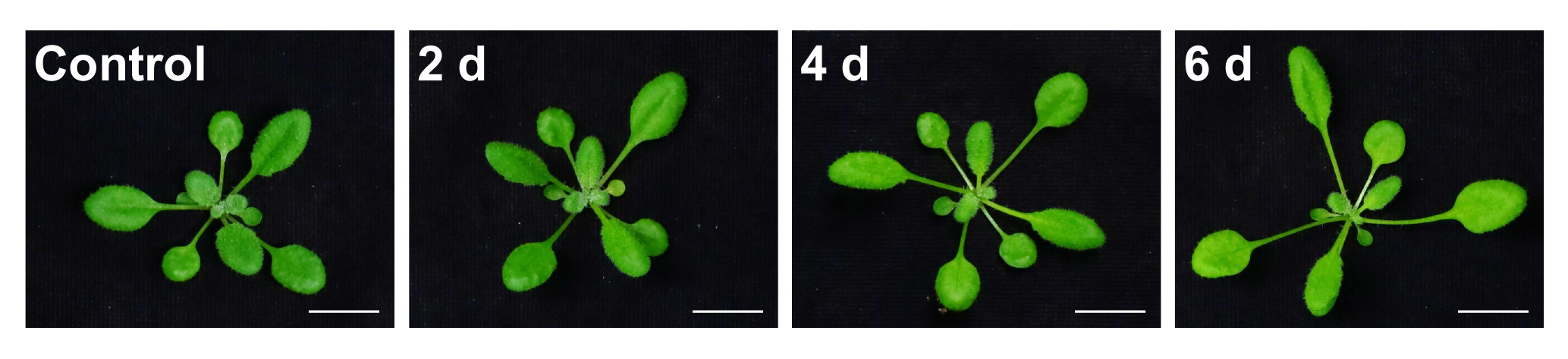} 
	\caption[Phenotype of wildtype Arabidopsis under high-temperature stress.]{\textbf{Phenotype of wildtype Arabidopsis exposed to various durations of high-temperature stress.} Wildtype Arabidopsis were grown at 21 \textsuperscript{o}C (control) before exposure to high temperature (29 \textsuperscript{o}C) over various durations---2 days (2d), 4 days (4d), and 6 days (6d). All plants were 3 w old at the end of the treatment, as illustrated in the previous supplementary figure. Scale, 1 cm.} 
	\label{SFig10}      
\end{figure}

\begin{figure}[htbp]
	\centering
	\includegraphics[width=\textwidth]{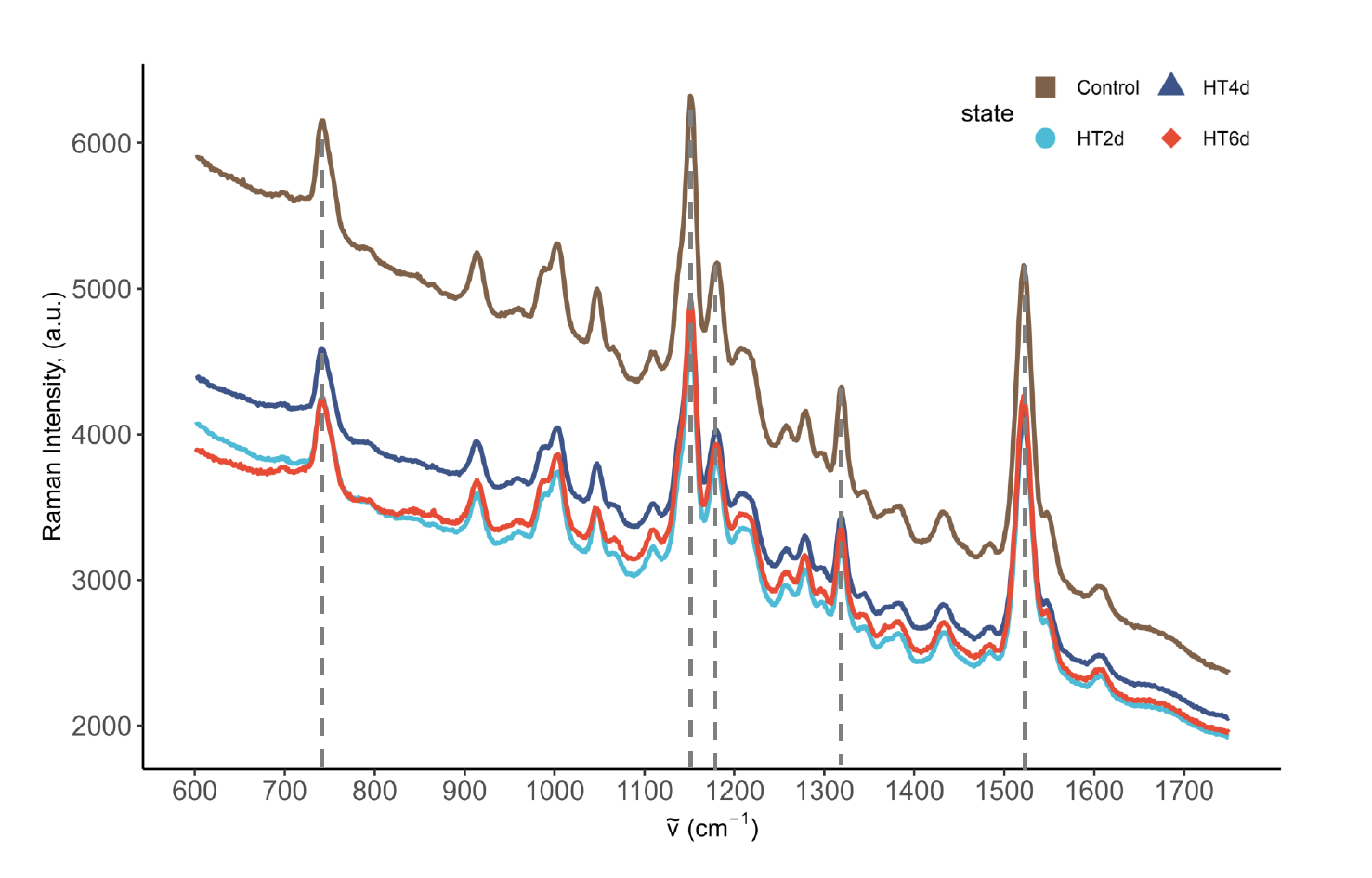} 
	\caption[Visualization of the original Raman spectra representing the temperature stress conditions in wildtype Arabidopsis leaves.]{\textbf{Visualization of the original Raman spectra representing the temperature stress conditions in wildtype Arabidopsis leaves.} The original spectra were obtained by detransforming the derivative D($\tilde{v}$) spectra reconstructed from the cluster medians of wildtype Arabidopsis leaf data in Fig.~4a. The five most significant peaks for each stress condition are highlighted with dotted lines. Control: 21 \textsuperscript{o}C, HT2d: Day-2 time point at high temperature (29 \textsuperscript{o}C), HT4d: Day-4 time point at high temperature (29 \textsuperscript{o}C), and HT6d: Day-6 time point at high temperature (29 \textsuperscript{o}C).} 
	\label{SFig11}      
\end{figure}
\clearpage

\begin{figure}[htbp]
	\centering
    \includegraphics[width=\textwidth]{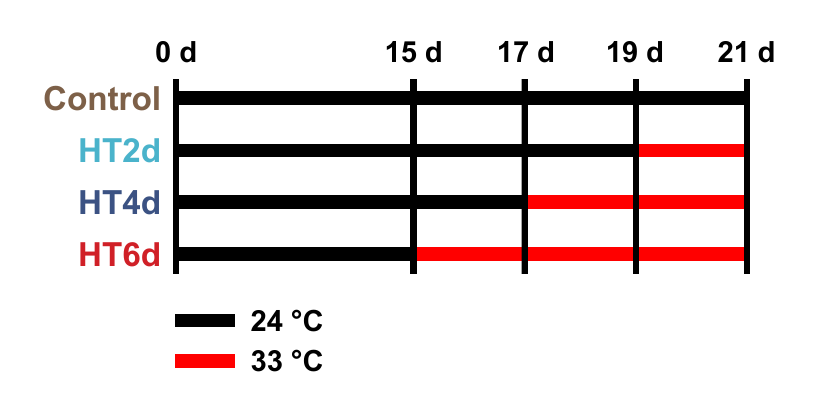} 
	\caption[Schematic diagram illustrating different durations of high-temperature treatment used for wildtype Choy Sum and Kai Lan.]{\textbf{Schematic diagram illustrating different durations of high-temperature treatment used for wildtype Choy Sum and Kai Lan.} Wildtype Choy Sum and Kai Lan were grown at 24 \textsuperscript{o}C before exposure to high temperature (33 \textsuperscript{o}C) over various durations. Control: 24 \textsuperscript{o}C, HT2d: Day-2 time point at high temperature (33 \textsuperscript{o}C), HT4d: Day-4 time point at high temperature (33 \textsuperscript{o}C), and HT6d: Day-6 time point at high temperature (33 \textsuperscript{o}C). All plants were 3 w old at the end of the treatment.} 
	\label{SFig12}      
\end{figure}
\clearpage

\begin{figure}[htbp]
	\centering
    \includegraphics[width=\textwidth]{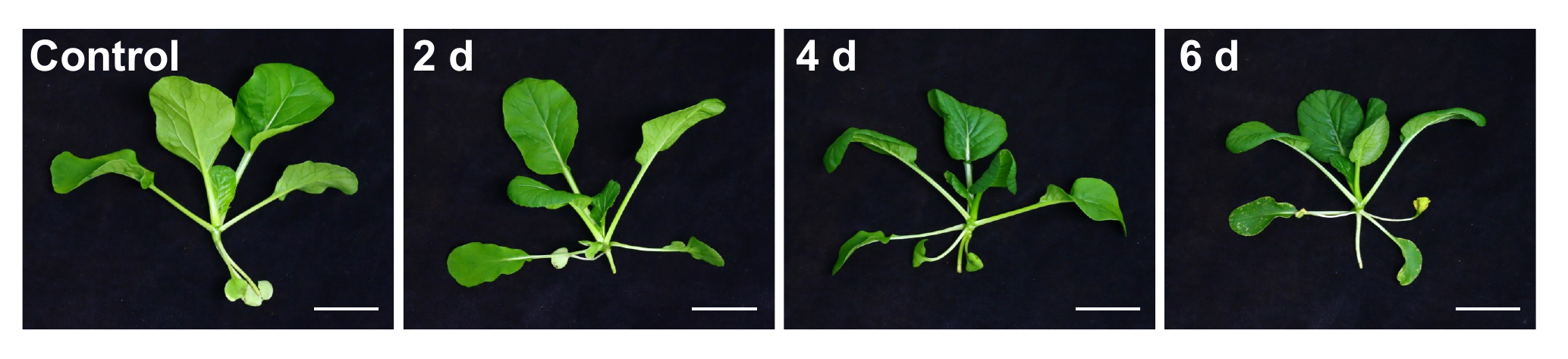} 
	\caption[Phenotypes of wildtype Choy Sum under high-temperature stress.]{\textbf{Phenotypes of wildtype Choy Sum exposed to various durations of high-temperature stress.} Wildtype Choy Sum were grown at 24 \textsuperscript{o}C before exposure to high temperature (33 \textsuperscript{o}C) over various durations---2 days (2d), 4 days (4d), and 6 days (6d). All plants were 3 w old at the end of the treatment, as illustrated in the previous supplementary figure. Scale, 3 cm.} 
	\label{SFig13}      
\end{figure}
\clearpage

\begin{figure}[htbp]
	\centering
	\includegraphics[width=\textwidth]{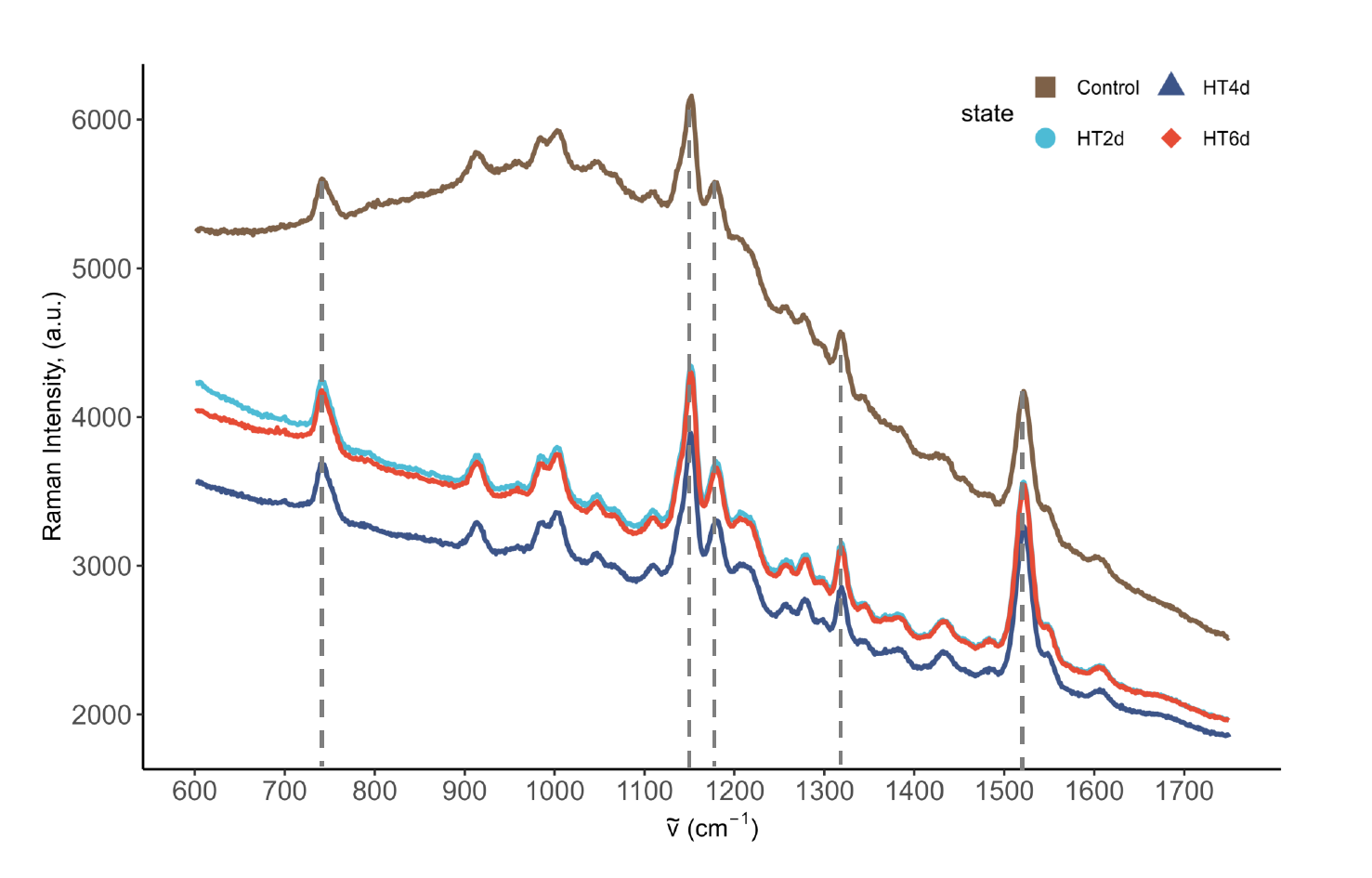} 
	\caption[Visualization of the original Raman spectra representing the temperature stress conditions in wildtype Choy Sum leaves.]{\textbf{Visualization of the original Raman spectra representing the temperature stress conditions in wildtype Choy Sum leaves.} The original spectra were obtained by detransforming the derivative D($\tilde{v}$) spectra reconstructed from the cluster medians of wildtype Choy Sum leaf data in Fig.~4c. The five most significant peaks for each stress condition are highlighted with dotted lines. Control: 24 \textsuperscript{o}C, HT2d: Day-2 time point at high temperature (33 \textsuperscript{o}C), HT4d: Day-4 time point at high temperature (33 \textsuperscript{o}C), and HT6d: Day-6 time point at high temperature (33 \textsuperscript{o}C).} 
	\label{SFig14}      
\end{figure}
\clearpage

\begin{figure}[htbp]
	\centering
    \includegraphics[width=\textwidth]{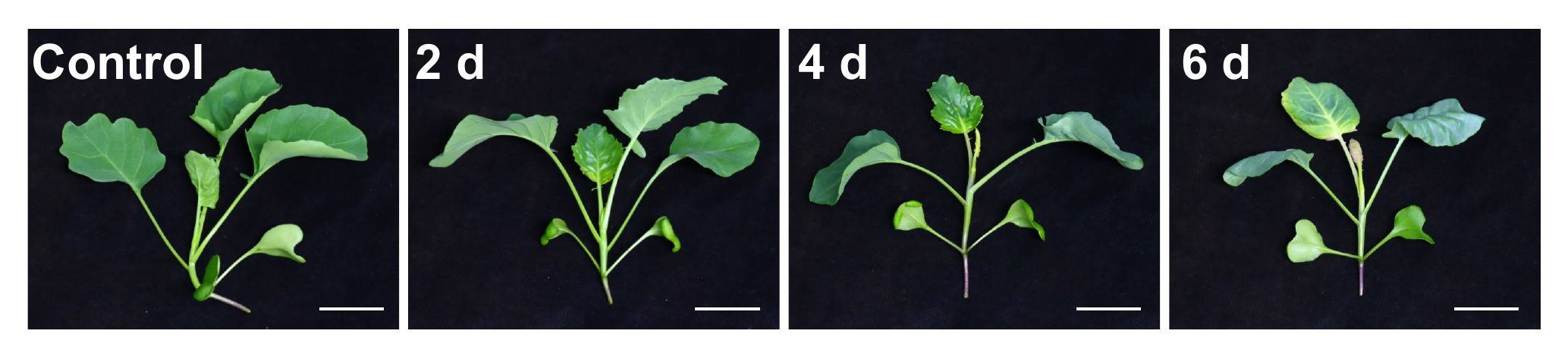} 
	\caption[Phenotypes of wildtype Kai Lan under high-temperature stress.]{\textbf{Phenotypes of wildtype Kai Lan exposed to various durations of high-temperature stress.} Wildtype Kai Lan were grown at 24 \textsuperscript{o}C before exposure to high ambient temperature (33 \textsuperscript{o}C) over various durations. All plants were 3 w old at the end of the treatment. Scale, 3 cm.} 
	\label{SFig15}      
\end{figure}
\clearpage

\begin{figure}[htbp]
	\centering
	\includegraphics[width=\textwidth]{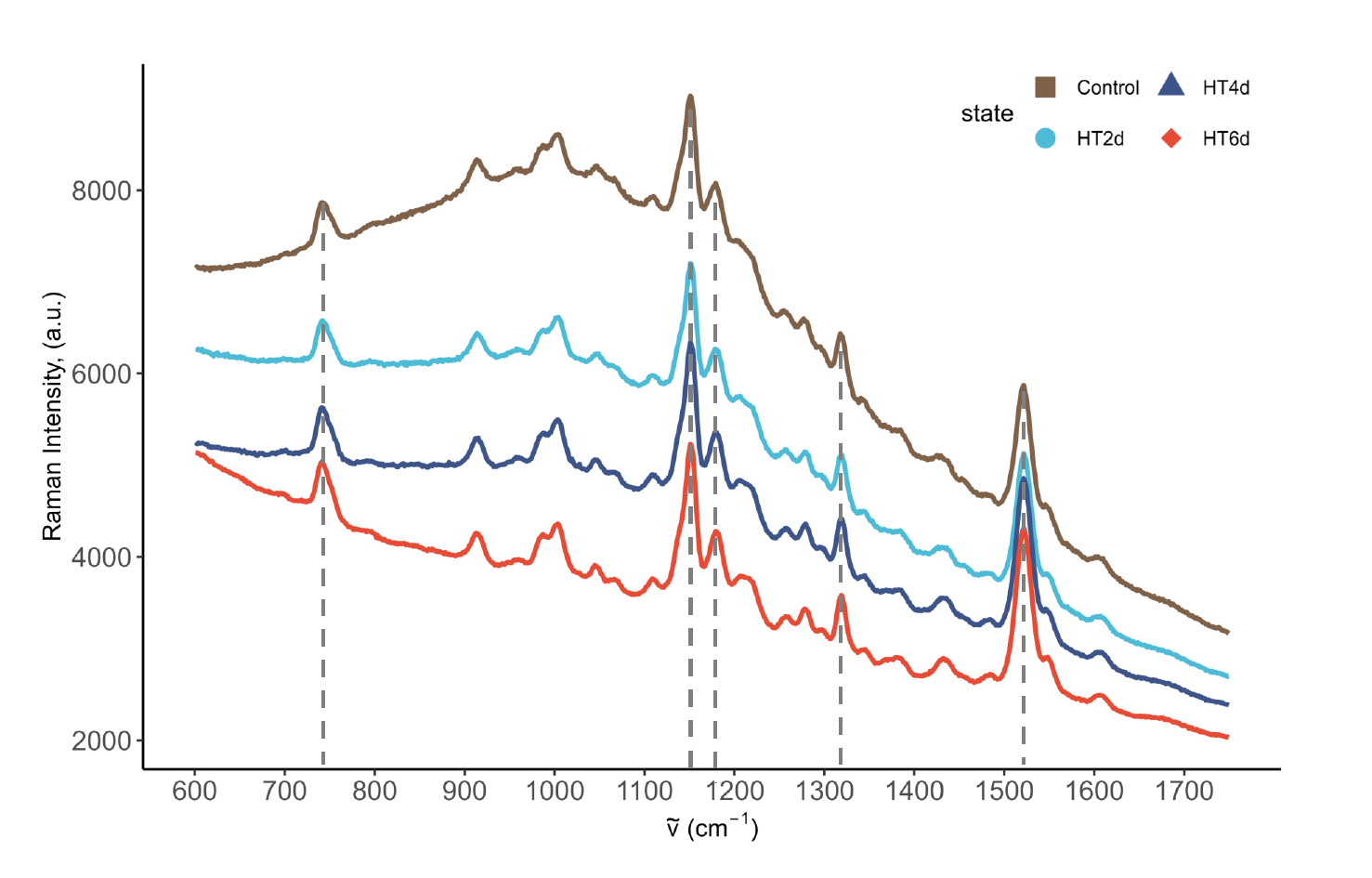} 
	\caption[Visualization of the original Raman spectra representing the temperature stress conditions in wildtype Kai Lan leaves.]{\textbf{Visualization of the original Raman spectra representing the temperature stress conditions in wildtype Kai Lan leaves.} The original spectra were obtained by detransforming the derivative D($\tilde{v}$) spectra reconstructed from the cluster medians of wildtype Kai Lan leaf data in Fig.~4e. The five most significant peaks for each stress condition are highlighted with dotted lines. Control: 24 \textsuperscript{o}C, HT2d: Day-2 time point at high temperature (33 \textsuperscript{o}C), HT4d: Day-4 time point at high temperature (33 \textsuperscript{o}C), and HT6d: Day-6 time point at high temperature (33 \textsuperscript{o}C).} 
	\label{SFig16}      
\end{figure}
\clearpage

\begin{figure}[htbp]
	\centering
    \includegraphics[width=\textwidth]{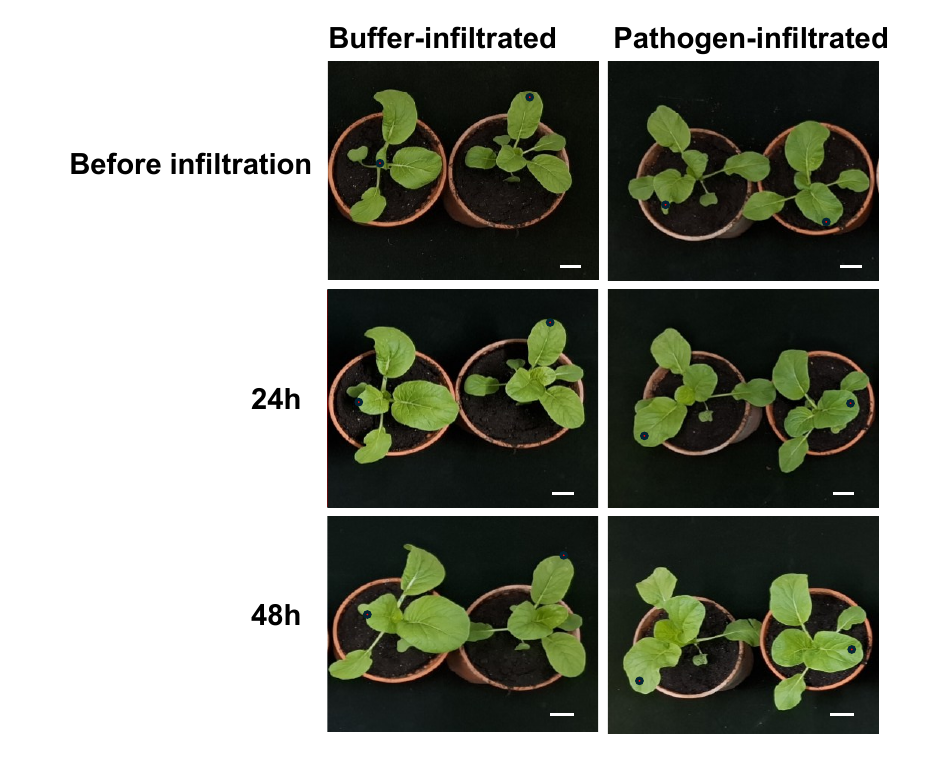} 
	\caption[Phenotypes of wildtype Choy Sum plants following infiltration with buffer and pathogenic bacterial suspension.]{\textbf{Phenotypes of wildtype Choy Sum plants following infiltration with buffer and bacterial suspension.} Wildtype Choy Sum plants were grown at 25 \textsuperscript{o}C before being infiltrated with buffer or bacterial suspensions. All plants were 2 w old at the time of infiltration. Both buffer- and pathogen-infiltrated leaves appeared healthy with no visible signs of infection at 48 hours post-infiltration. Scale, 5 cm.} 
	\label{SFig17}      
\end{figure}
\clearpage

\begin{figure}[htbp]
	\centering
	\includegraphics[width=\textwidth]{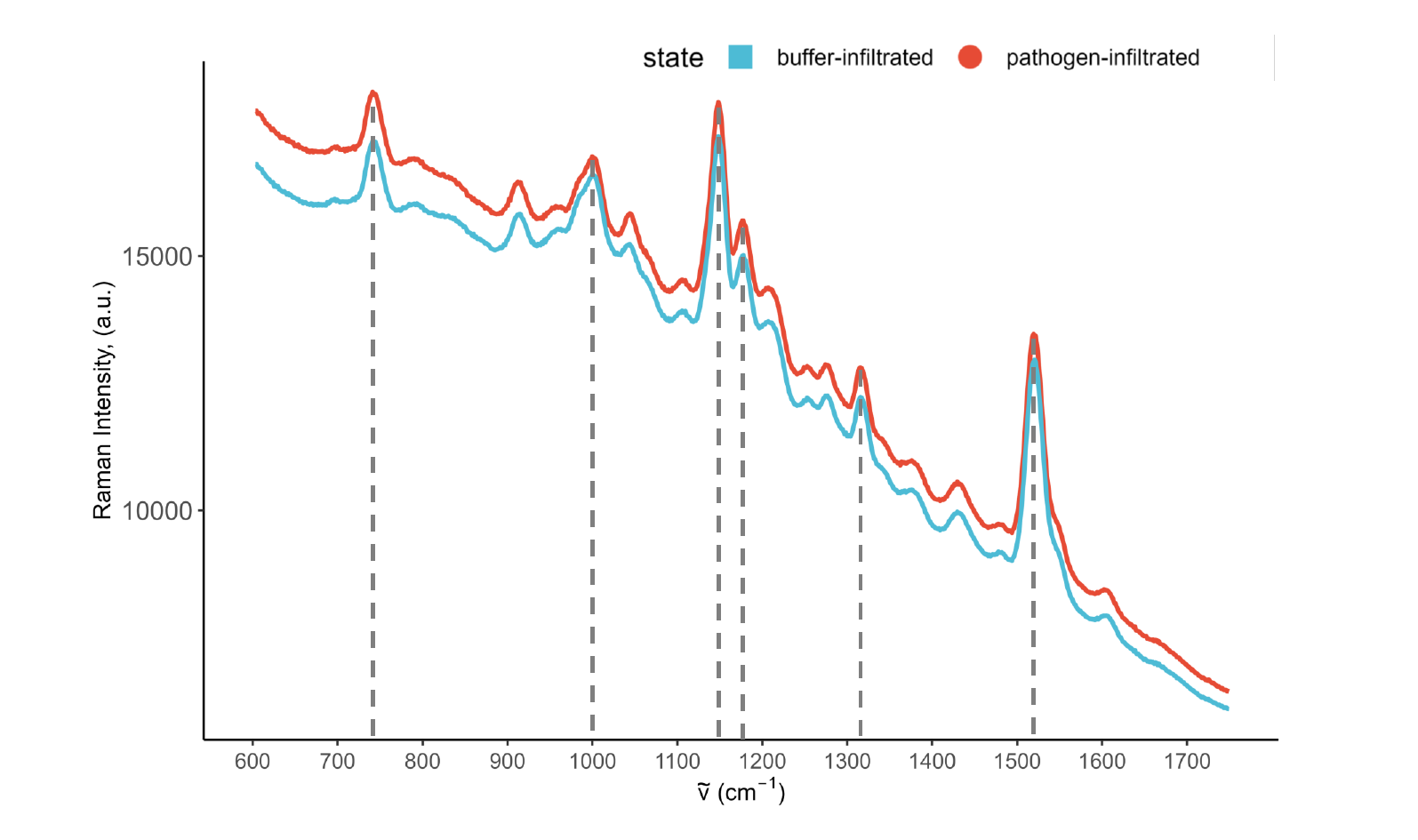} 
	\caption[Visualization of the original Raman spectra representing bacterial stress conditions at the 24-hour time point post-infection in wildtype Choy Sum leaves.]{\textbf{Visualization of the original Raman spectra representing bacterial stress conditions at the 24-hour time point post-infection in wild-type Choy Sum leaves.} The original spectra were obtained by detransforming the derivative D($\tilde{v}$) spectra reconstructed from the cluster medians of wildtype Choy Sum leaf data in Fig.~5a. The most significant peaks for the buffer-infiltrated and pathogen-infiltrated plant groups are highlighted with dotted lines.} 
	\label{SFig18}      
\end{figure}
\clearpage

\begin{figure}[htbp]
	\centering
	\includegraphics[width=\textwidth]{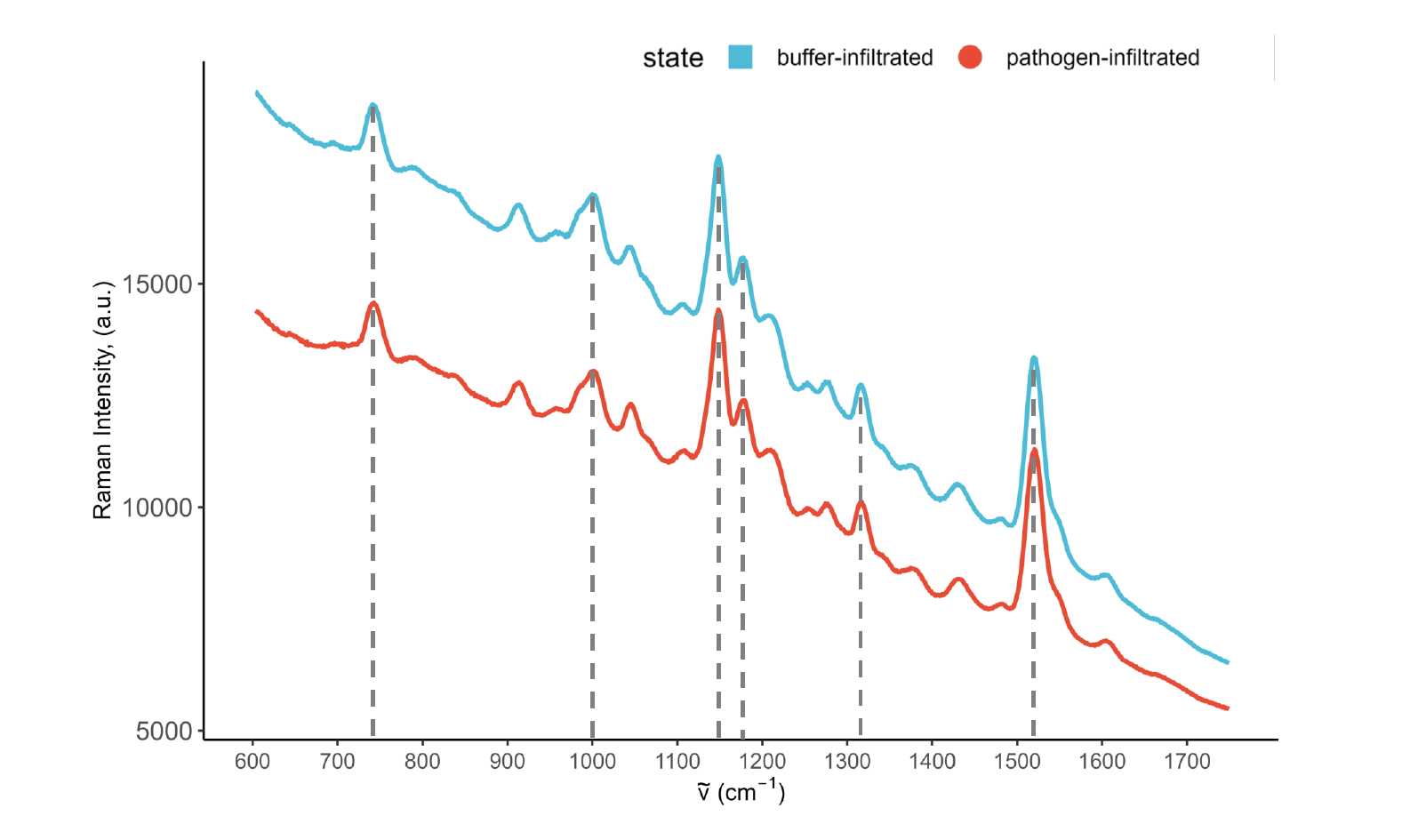} 
	\caption[Visualization of the original Raman spectra representing bacterial stress conditions at the 48-hour time point post-infection in wildtype Choy Sum leaves.]{\textbf{Visualization of the original Raman spectra representing bacterial stress conditions at the 48-hour time point post-infection in wild-type Choy Sum leaves.} The original spectra were obtained by detransforming the derivative D($\tilde{v}$) spectra reconstructed from the cluster medians of wildtype Choy Sum leaf data in Fig.~5c. The most significant peaks for the buffer-infiltrated and pathogen-infiltrated plant groups are highlighted with dotted lines.} 
	\label{SFig19}      
\end{figure}
\clearpage

\begin{figure}[htbp]
	\centering
	\includegraphics[width=\textwidth]{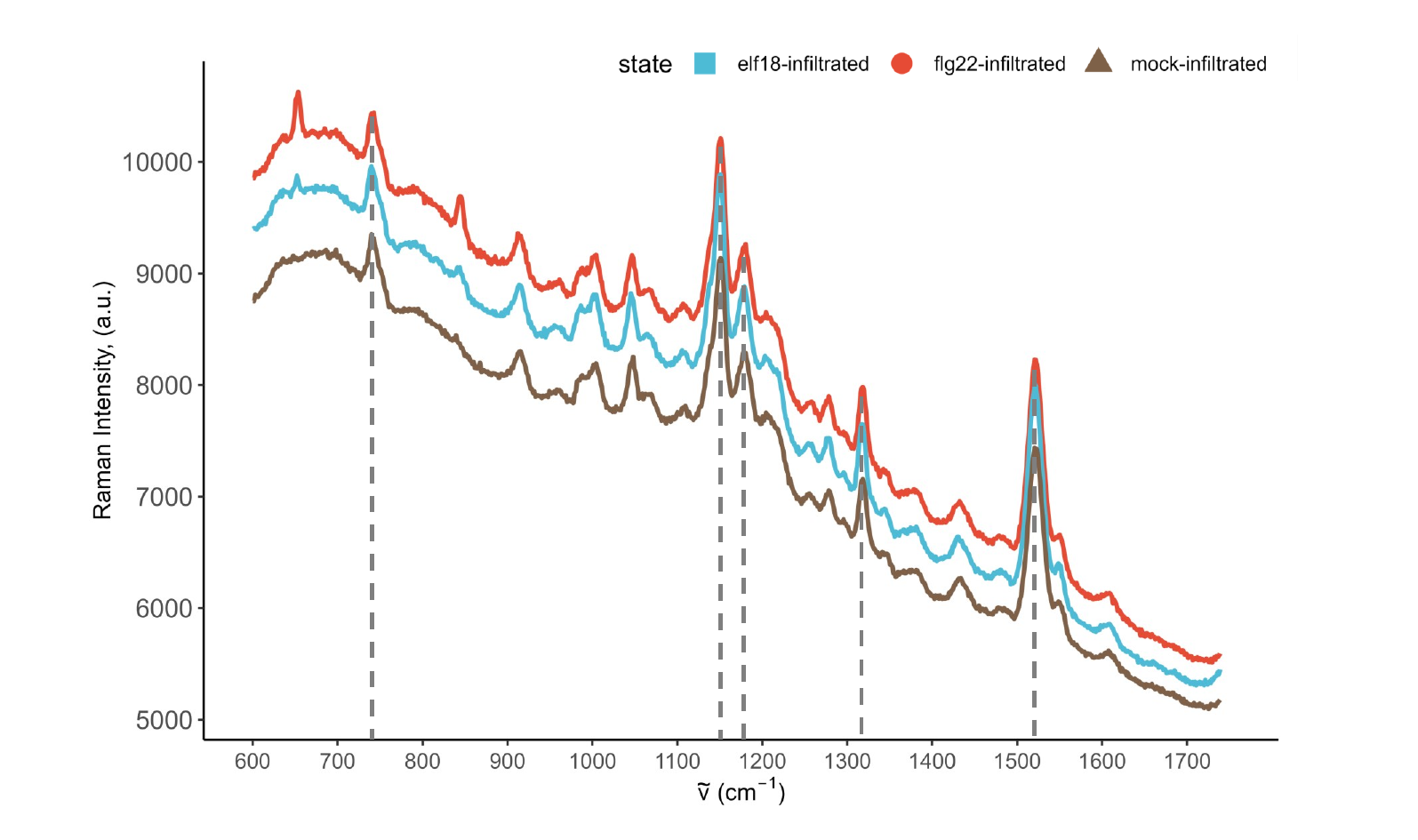} 
	\caption[Visualization of the original Raman spectra representing stress conditions induced by bacterial elicitors at the 24-hour time point post-infection in wildtype Arabidopsis leaves.]{\textbf{Visualization of the original Raman spectra representing stress conditions induced by bacterial elicitors at the 24-hour time point post-infection in wildtype Arabidopsis leaves.} The original spectra were obtained by detransforming the derivative D($\tilde{v}$) spectra reconstructed from the cluster medians of wildtype Arabidopsis leaf data in Extended~Fig.~1a. The most significant peaks for the mock-infiltrated, elf18-infiltrated and flg22-infiltrated plant groups are highlighted with dotted lines.} 
	\label{SFig20}      
\end{figure}
\clearpage

\begin{figure}[htbp]
	\centering
	\includegraphics[width=\textwidth]{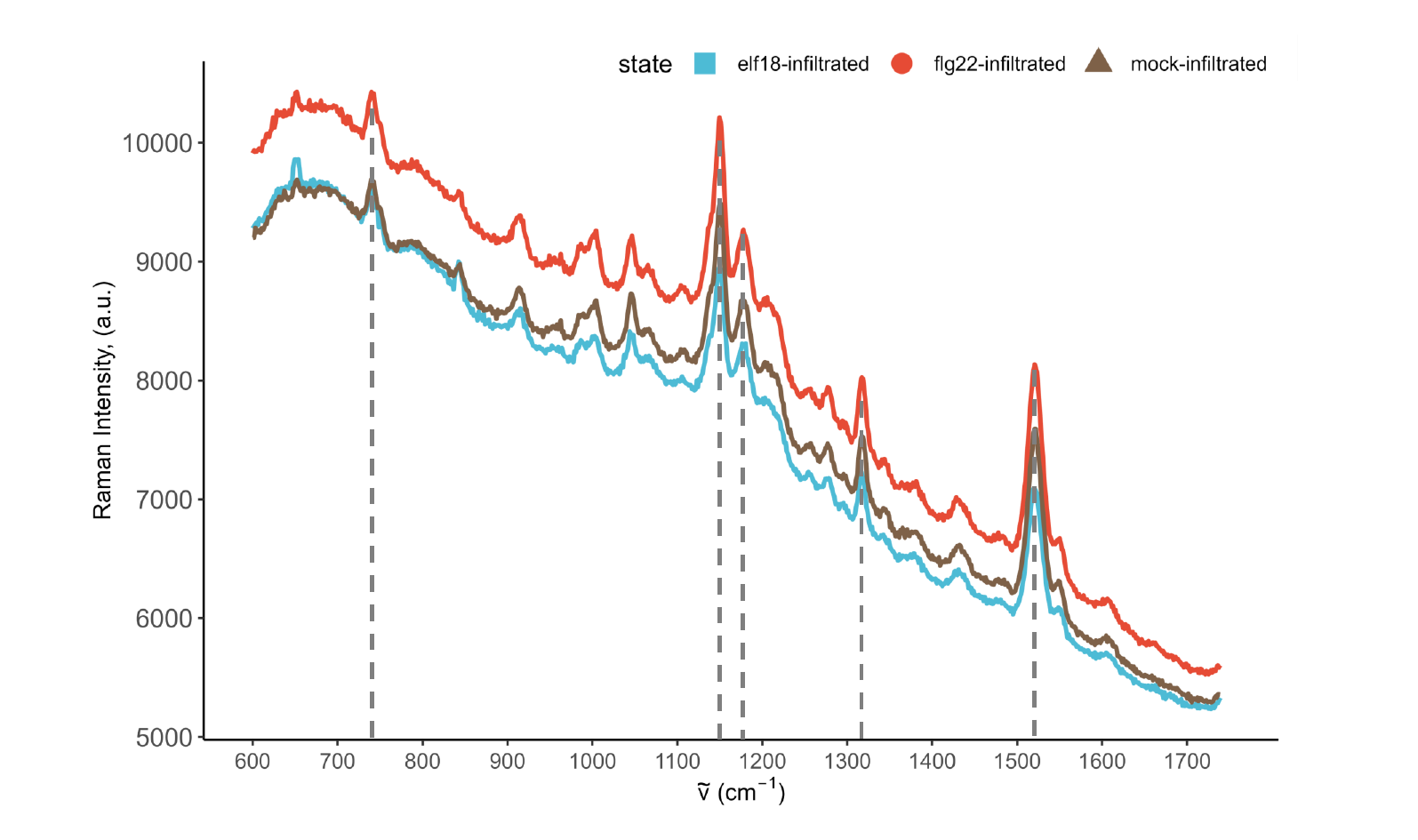} 
	\caption[Visualization of the original Raman spectra representing stress conditions induced by bacterial elicitors at the 48-hour time point post-infection in wildtype Arabidopsis leaves.]{\textbf{Visualization of the original Raman spectra representing stress conditions induced by bacterial elicitors at the 48-hour time point post-infection in wildtype Arabidopsis leaves.} The original spectra were obtained by detransforming the derivative D($\tilde{v}$) spectra reconstructed from the cluster medians of wildtype Arabidopsis leaf data in Extended~Fig.~1c. The most significant peaks for the mock-infiltrated, elf18-infiltrated and flg22-infiltrated plant groups are highlighted with dotted lines.} 
	\label{SFig21}      
\end{figure}
\clearpage

\begin{figure}[htbp]
	\centering
	\includegraphics[width=\textwidth]{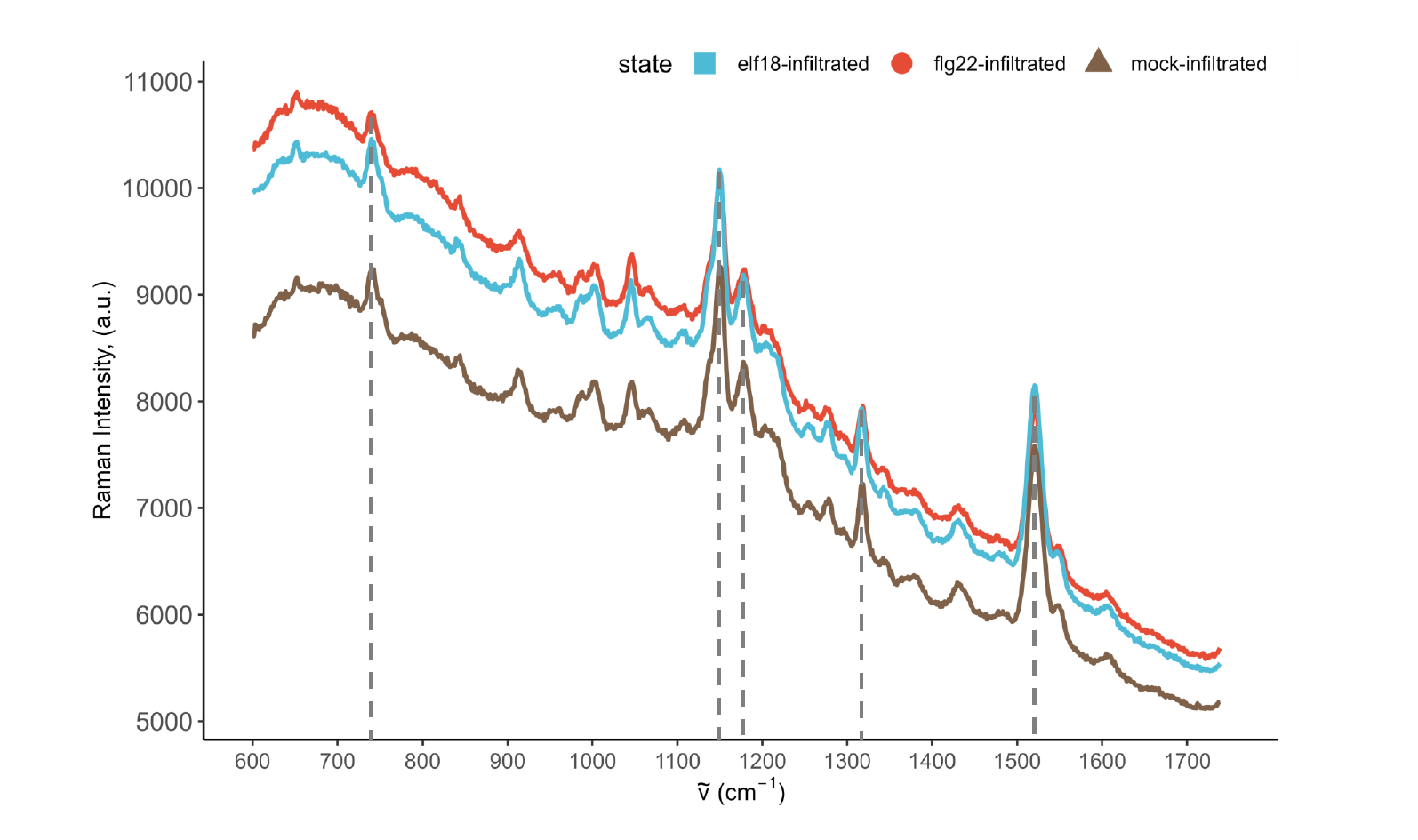} 
	\caption[Visualization of the original Raman spectra representing stress conditions induced by bacterial elicitors at the 72-hour time point post-infection in wildtype Arabidopsis leaves.]{\textbf{Visualization of the original Raman spectra representing stress conditions induced by bacterial elicitors at the 72-hour time point post-infection in wildtype Arabidopsis leaves.} The original spectra were obtained by detransforming the derivative D($\tilde{v}$) spectra reconstructed from the cluster medians of wildtype Arabidopsis leaf data in Extended~Fig.~1e. The most significant peaks for the mock-infiltrated, elf18-infiltrated and flg22-infiltrated plant groups are highlighted with dotted lines.} 
	\label{SFig22}      
\end{figure}
\clearpage

\section{Supplementary Tables}

\begin{table}[h]
\caption[Summary of biomolecular changes in response to light stress across various light conditions in Arabidopsis, Choy Sum, and Kai Lan.]{Summary of biomolecular changes in response to light stress across various light conditions (high light, white light, low light, and shade) in Arabidopsis, Choy Sum, and Kai Lan. The table highlights the changes in key biomolecules—carotenoids, cellulose, lignin, proteins, and pectin—reflecting the plant's adaptive responses to varying light environments.}
\label{STab1}
\vspace{5mm}
\begin{tabular}{|l|l|l|l|}
\hline
\multicolumn{1}{|c|}{\textbf{Light Condition}} & \multicolumn{1}{c|}{\textbf{Cellulose/Lignin}} & \multicolumn{1}{c|}{\textbf{Protein}} & \multicolumn{1}{c|}{\textbf{Pectin}} \\ \hline
High Light & ↑ & ↑ & ↑ \\ \hline
White Light & — (baseline) & — & — \\ \hline
Low Light & ↓ & ↓ & ↓ / $<$---$>$ (shift) \\ \hline
Shade & ↓  & ↓ / $<$---$>$ & ↓ /$<$---$>$ \\ \hline
\end{tabular}
\end{table}